\documentclass[10pt,twocolumn,letterpaper]{article}

\usepackage[pagenumbers]{iccv} 

\usepackage{booktabs}
\usepackage{arydshln}
\usepackage{bbding}
\usepackage{mathtools}
\usepackage{multirow}

\definecolor{iccvblue}{rgb}{0.21,0.49,0.74}
\usepackage[pagebackref,breaklinks,colorlinks,allcolors=iccvblue]{hyperref}

\def\paperID{9078} 
\def\confName{ICCV}
\def\confYear{2025}

\title{Ensemble Foreground Management for Unsupervised Object Discovery\thanks{\tiny Author’s Accepted Manuscript. Released under the Creative Commons license: Attribution 4.0 International (CC BY 4.0) \url{https://creativecommons.org/licenses/by/4.0/}}}

\author{Ziling Wu\thanks{\tiny Corresponding author}, Armaghan Moemeni, Praminda Caleb-Solly\\
School of Computer Science, University of Nottingham, UK\\
{\tt\small \{ziling.wu, armaghan.moemeni, praminda.caleb-solly\}@nottingham.ac.uk}
}

\begin{document}
\maketitle
\begin{abstract}
Unsupervised object discovery (UOD) aims to detect and segment objects in 2D images without handcrafted annotations. Recent progress in self-supervised representation learning~\cite{denseCL, dino} has led to some success in UOD algorithms~\cite{simeoni2021localizing,wang2022tokencut,wang2023cut}. However, the absence of ground truth provides existing UOD methods with two challenges: 1) determining if a discovered region is foreground or background, and 2) knowing how many objects remain undiscovered. To address these two problems, previous solutions rely on foreground priors~\cite{simeoni2021localizing, wang2023cut, MaskDistill, freesolo} to distinguish if the discovered region is foreground, and conduct one or fixed iterations of discovery. However, the existing foreground priors are heuristic and not always robust, and a fixed number of discoveries leads to under or over-segmentation, since the number of objects in images varies. This paper introduces UnionCut, a robust and well-grounded foreground prior based on min-cut~\cite{max_flow} and ensemble methods~\cite{ensemble} that detects the union of foreground areas of an image, allowing UOD algorithms to identify foreground objects and stop discovery once the majority of the foreground union in the image is segmented. In addition, we propose UnionSeg, a distilled transformer of UnionCut that outputs the foreground union more efficiently and accurately. Our experiments show that by combining with UnionCut or UnionSeg, previous state-of-the-art UOD methods~\cite{wang2023cut, wang2022tokencut, FOUND, simeoni2021localizing} witness an increase in the performance of single object discovery, saliency detection and self-supervised instance segmentation on various benchmarks. The code is available at \url{https://github.com/YFaris/UnionCut}.
\end{abstract}
\section{Introduction}
\label{sec:intro}
Instance segmentation is a fundamental computer vision task requiring the algorithm to recognize and segment each instance in an image, which has been widely applied in autonomous driving~\cite{cityscapes}, medical image analysis~\cite{unet}, remote sensing~\cite{remote_sensing}, etc. However, pixel-level annotations are required to train an instance segmentation model, which are expensive and time-consuming to obtain. To cut down the training cost of instance segmentation models, strategies employing semi-supervision~\cite{119,cct,cps,mean_teacher,bb-unet,sam,semi_gan1, semi_gan2}, weak supervision~\cite{Box2seg, Boxinst, bbam,pcams, point2,image1,image2,image3}, and self-supervision~\cite{freesolo,simeoni2021localizing,wang2022tokencut,wang2023cut, ProMerge} have been proposed to reduce the usage of pixel-level annotations. 
    
\begin{figure}[t]
    \centering
    \includegraphics[width=8.3cm]{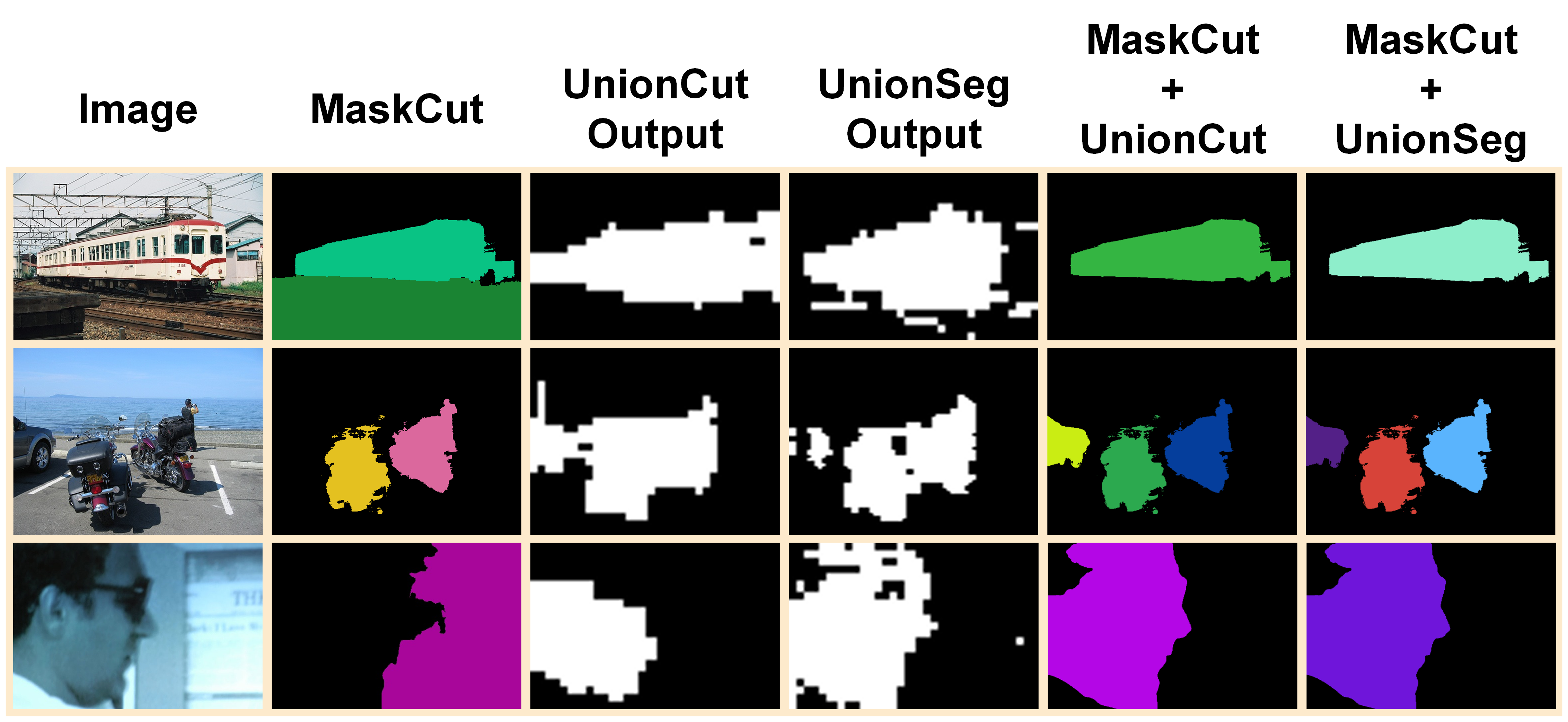}
    \caption{Examples of fixing MaskCut~\cite{wang2023cut}\textquotesingle s errors and making its discovery stop at appropriate time using UnionCut and UnionSeg.}
    \label{intro_demo}
\end{figure}

Self-supervision aims to train an instance segmentation model without any annotations. It utilizes an unsupervised object discovery (UOD) algorithm to detect and segment objects in images without handcrafted annotations, following which, the detection result is used as the ground truth to train a segmentation model~\cite{force1, force2, force3, freesolo, wang2022tokencut, ProMerge}. Recent state-of-the-art UOD algorithms~\cite{simeoni2021localizing,wang2022tokencut,wang2023cut, freesolo, ProMerge} are based on a vision transformer (ViT)~\cite{vit} pre-trained by self-supervised representation learning~\cite{dino}. These draw on a phenomenon where the attention map of the last layer of the ViT indicates regions of objects in the image. However, without the foreground knowledge given by the ground truth, existing UOD methods cannot robustly judge whether the explored region belongs to the foreground or not, and as such, the algorithm could return the background area of the image as a discovered object. As a remedy, existing approaches introduce foreground priors (\eg the location and size of the discovered region) to help them distinguish if an area in the image is foreground. However, we have found that existing foreground priors are not robust because they are based on simple rules, assumptions, and heuristic design. Besides, the number of objects in the image varies and is unknown to UOD approaches, resulting in a dilemma of when to stop the discovery. Thus, existing UOD algorithms either segment only one region as the object or continue discovery for a predetermined number of iterations, leading to either missing objects or mistakenly identifying background as foreground due to unreliable foreground priors and excessive exploration. For example, in \cref{intro_demo}, MaskCut~\cite{wang2023cut}, a UOD method, returns a background area (the 1st and 3rd row), and misses the car on the left of the 2nd image.

To address the problem, we propose an ensemble method named UnionCut. It detects the union of foreground regions in an image, referred to as ``foreground union" in this paper. With this foreground union as the foreground prior, UOD methods can accurately judge if a discovered region is the foreground by checking if the foreground union covers it, and stop further exploration when the majority of the foreground union has been discovered, as shown in~\cref{intro_demo}. Unlike previous assumption-based foreground priors, UnionCut\textquotesingle s robustness and reliability are demonstrated with mathematical and statistical ensemble theories. Besides, we propose UnionSeg, a ViT distilled from UnionCut to increase the efficiency and accuracy of foreground union detection.

Our work makes the following contributions: \textbf{\textit{Insights:}} Previous UOD studies focused mainly on the design of UOD algorithms, overlooking foreground priors\textquotesingle~importance. We reveal that UOD\textquotesingle s performance can also be increased by robust foreground priors. \textbf{\textit{Methods:}} We propose UnionCut based on ensemble and UnionSeg using distillation to detect foreground union in the image, serving as foreground priors for UOD methods to verify results and decide when to stop discovery. Unlike previous heuristic and assumption-based foreground priors, UnionCut is mathematically and statistically robust and reliable. \textbf{\textit{Results:}} After being integrated with UnionCut or UnionSeg, previous state-of-the-art UOD methods~\cite{wang2023cut, wang2022tokencut, FOUND, simeoni2021localizing} witness an increase in the performance of single object discovery, saliency detection and self-supervised instance segmentation on various benchmarks, indicating that UnionCut and UnionSeg enable UOD algorithms to give more accurate predictions on the existence of objects in the image. \textbf{\textit{Impacts:}} UnionCut and UnionSeg can be used as future UOD methods\textquotesingle~default foreground priors, taking the place of existing assumption-based, heuristic, and unreliable foreground priors, and enhancing UOD performance.

\section{Related Work}
\label{sec:review}
\subsection{Self-supervised Representation Learning}
Self-supervised representation learning aims to train the model to generate distinguishable embeddings for images and output similar embeddings of images with the same semantic information without using handcrafted annotations. Most existing approaches~\cite{moco,byol,dino,dinov2,ibot,denseCL,swav,simsiam,simclr} use a student-teacher architecture, where each model processes one of two differently augmented versions (\eg horizontal flipping, randomly cropping, colour jitter, etc.) of the same image. The optimization goal is to make the models produce similar outputs for augmentations of the same image and distinct outputs for augmentations from different images. DINO~\cite{dino} is a representative method of these approaches, where the attention map of the last layer of a ViT pre-trained by DINO indicates the location of objects in the image. Such a property is employed in our approach. 

\begin{figure*}[t]
    \centering
    \includegraphics[height=5.3cm]{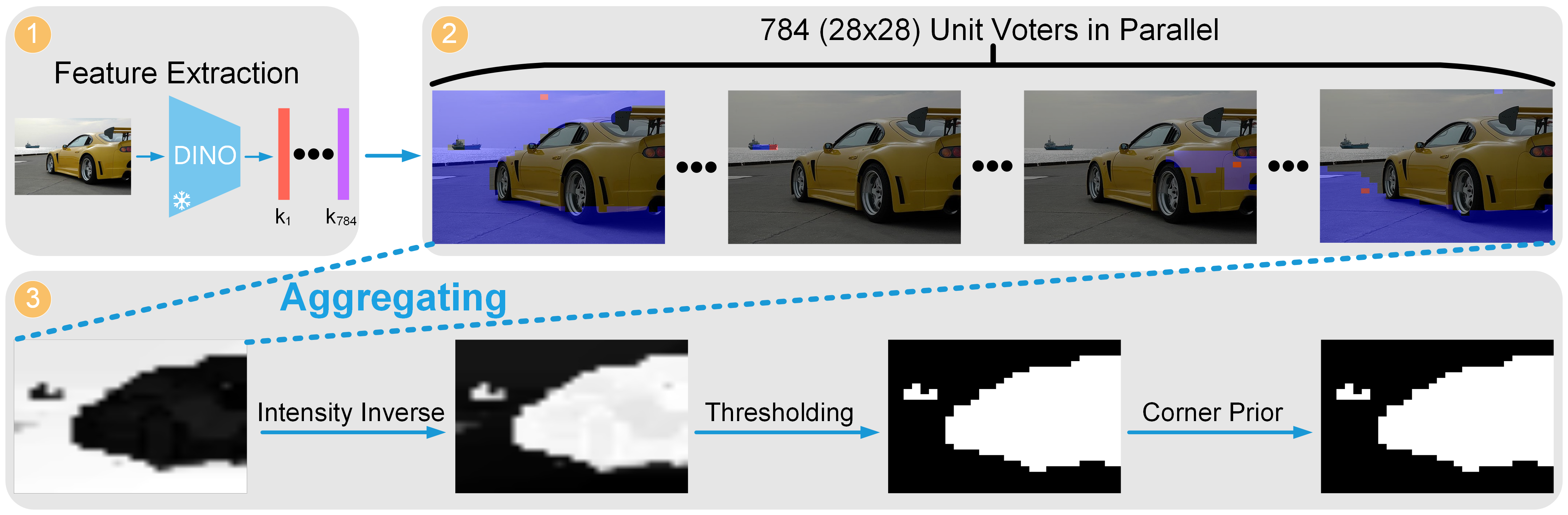}
    \caption{\textbf{Framework of UnionCut}. UnionCut takes an image as input with features extracted from a frozen DINO, and applies Unit Voters (UVs) with each using one patch as its seed. Red rectangles refer to the seed patch, and blue areas indicate the UV\textquotesingle s output of the region similar to its seed patch. Aggregating all UV\textquotesingle s outputs yields a heat map of the background, which is then inverted, thresholded, and rectified with a corner prior to produce a binary mask of the image\textquotesingle s foreground union.}
    \label{unioncut_framework}
\end{figure*}

\subsection{Unsupervised Object Discovery}
As for UOD, Wang \etal~\cite{freesolo} proposed FreeMask to discover objects by comparing the similarity between features from a self-supervised CNN~\cite{denseCL} at each pixel of the image. Instead of using CNN, in \cite{simeoni2021localizing,MaskDistill,wang2022tokencut,wang2023cut}, embeddings of patches output by a ViT pre-trained by DINO\cite{dino} are utilized to generate pseudo-masks. In \cite{simeoni2021localizing, MaskDistill}, a patch is selected as the seed, and the pseudo mask is given by combining patches sharing similar features with the seed. To increase the pseudo-mask quality, Normalized Cut (NCut) \cite{ncut} is leveraged by Wang~\etal~\cite{wang2022tokencut} to segment the feature map output by DINO for pseudo-mask generation. In \cite{wang2023cut}, NCut is expanded to segment multiple objects in an image by iteratively removing a new segmented region from the image. However, the absence of ground truth makes it hard to determine if any discovered region is actually part of the foreground. Our approach provides a solution to this. 

\subsection{Foreground Priors}
\label{literature_foreground_prior}
UOD algorithms leverage foreground priors to judge if the discovered region belongs to the foreground. Specifically, Siméoni \etal~\cite{simeoni2021localizing} assumes that the area occupied by the foreground should be smaller than the background. FreeMask~\cite{freesolo} removes discovered regions whose widths are longer than 95\% of the width of the image or are located in the top 20\% region of the image. Gansbeke \etal~\cite{MaskDistill} and Siméoni \etal~\cite{FOUND} proposed that the semantic information of the foreground should be close to the global semantic information of the entire image. In~\cite{wang2022tokencut, wang2023cut}, the region where the point with the highest absolute confidence value is located is made as the foreground. Wang \etal~\cite{wang2023cut} suggested that the foreground is unlikely to cover three or four corners of the image. Li and Shin~\cite{ProMerge} proposed that a background mask is likely to occupy at least two edges of the image. Although these priors achieved limited success, they are all heuristic and assumption-based. In contrast, the robustness and reliability of our proposed methods are demonstrated by mathematical and statistical ensemble theories.

\subsection{Ensemble Methods}
Ensemble methods are learning algorithms that construct a set of classifiers and then classify new data points by taking a (weighted)
vote of their predictions~\cite{ensemble}. Most ensemble methods can be classified into two types: 1) \textbf{Parallel} a set of weak classifiers run in parallel and form a strong classifier by performing a plurality vote, \eg Pasting~\cite{pasting}, Bagging~\cite{bagging}, and Random Forest~\cite{random_forest}; and 2) \textbf{Sequential} running batches of weak classifiers iteratively with each batch focusing on hard cases of the last one, \eg~\cite{Viola-Jones} and AdaBoost~\cite{adaboost}. UnionCut is of the former: a set of weak classifiers is created for an image, with each in parallel voting every patch ($8 \times 8$ pixels) of the image as foreground or background, which forms a strong classifier by aggregating votes from them to detect the foreground union of an image.
\section{Methodology}
UnionCut is an ensemble method, enabling UOD methods to judge if a detected region is foreground and stop discovery when most objects have been discovered to avoid over-segmentation. In the context of ensemble methods, UnionCut serves as the strong classifier, based on many parallel weak classifiers named ``unit voter" (UV) here. \cref{unioncut_framework} illustrates the frameworks of the proposed UnionCut. Unlike previous assumption-based foreground priors mentioned in~\cref{literature_foreground_prior}, UnionCut is reliable and robust with ensemble theories. First, we introduce the details of UV.

\subsection{Unit Voter}
UV is based on features output by a ViT pretrained by DINO~\cite{dino}. In our implementation, an image is resized into $224 \times 224$ and serialized into $28 \times 28$ (\ie 784) patches of $8 \times 8$ pixels before feature extraction of each patch with the ViT. The set of patches of an image is defined as $P = \{p_1, p_2,...,p_{784}\}$, with the corresponding L2-normalized Key vectors as their features output by the last layer of the model, which is defined as $K = \{k_1, k_2,...,k_{784}\}$. Then, the patches and features are leveraged by each UV.

\begin{figure}[t]
    \centering
    \includegraphics[width=5.0cm]{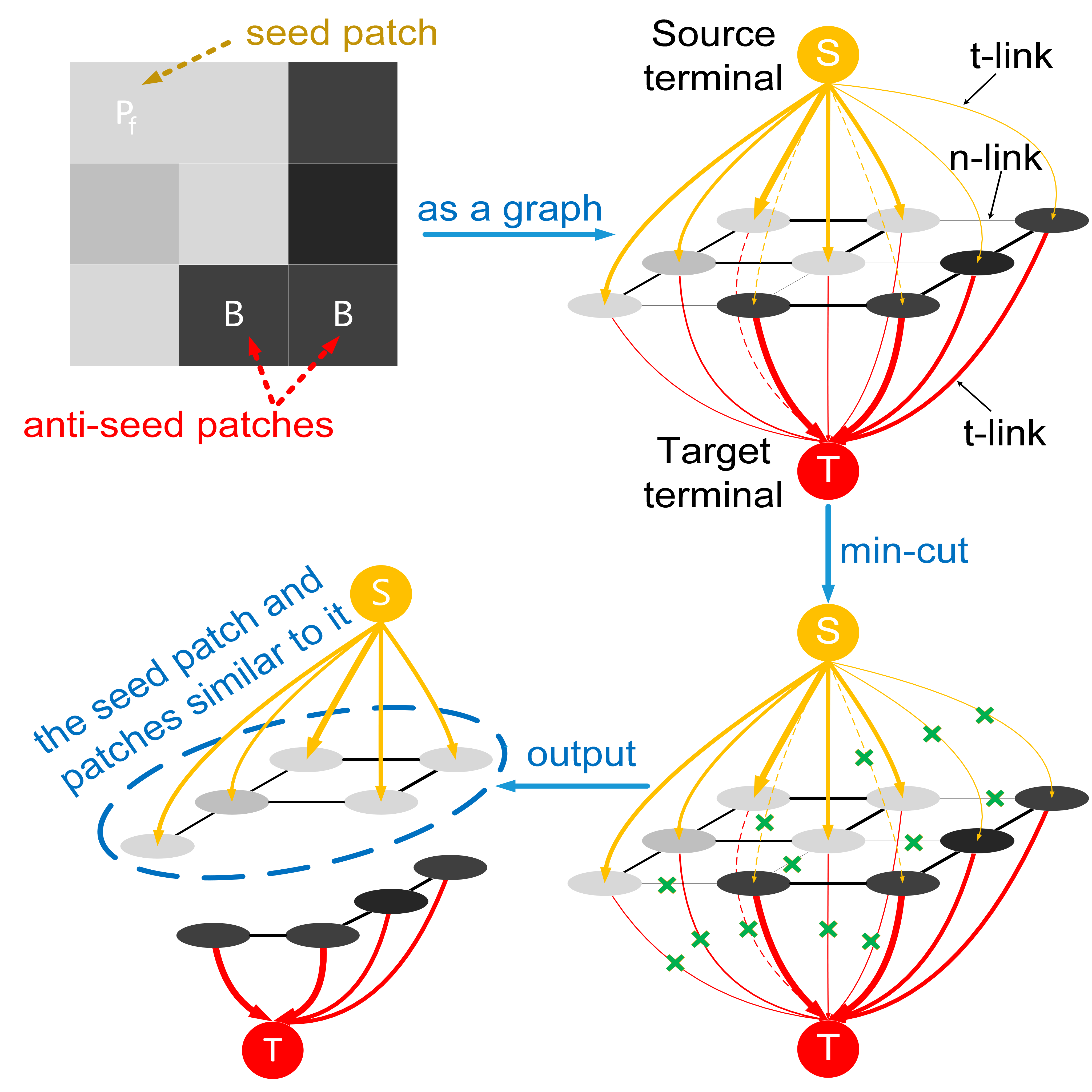}
    \caption{\textbf{Overview of Unit Voter (UV)}. For clarity, the image is divided into $3 \times 3$ patches. With the seed patch and anti-seed patches of the UV initialized, the image is modeled as a graph with each patch and two dummy terminals being nodes and connected by edges whose thickness reflects their weights. After applying min-cut to the graph, thin edges are removed to divide the graph into two disjoint parts containing one terminal node. The part with the Source node is the region output by the UV, indicating the area of features similar to the seed patch.}
    \label{graph_cut_intro}
\end{figure}

As required by ensemble methods, the number of weak classifiers (\ie UV) to form a strong classifier (\ie UnionCut) should be significant, and each UV should be different~\cite{ensemble}. To achieve these goals, 784 independent UVs can be built based on the 784 patches of an image, with each UV corresponding to a unique patch $p_f \in P$. The patch corresponding to a UV is called that UV\textquotesingle s ``seed patch". Taking a UV built from a seed patch $p_f \in P$ as an example, its pipeline is illustrated in ~\cref{graph_cut_intro}. Specifically, given a UV\textquotesingle s seed patch $p_f \in P$ with feature $k_f \in K$, the set of patches dissimilar to $p_f$, defined as the UV\textquotesingle s anti-seed patches set $B_f=\{p_b|p_b \in P, b \neq f, k_b^Tk_f<0 \}$, can be obtained.  With $p_f$, $B_f$, all $28 \times 28$ patches $P$ and features $K$, a directed graph is constructed, as shown in~\cref{graph_cut_intro}, where each patch is a node and two dummy terminal nodes, named ``Source" and ``Target", are added. In the graph, nodes of adjacent patches (\ie eight-neighborhood patches) are connected by bidirectional edges (named ``n-links") with weights representing their similarity. Directed edges named ``t-links" connect the Source to each patch node or each patch node to the Target, with weights based on the likelihood of a patch being more similar to the UV\textquotesingle s seed patch $p_f$ or its anti-seed patches set $B_f$. The terms ``t-link" and ``n-link" originate from early image segmentation methods that modeled images as graphs ~\cite{graph_cut, grabcut}. The purpose of building this graph is to split the image by removing edges, forming two disconnected subgraphs with each either containing the Source or the Target, while minimizing the total weight of the removed edges. The subgraph with the Source is expected to contain patches similar to the UV\textquotesingle s seed patch, while the one with the Target includes patches similar to the anti-seed patches $B_f$. This bipartition process is also known as the min-cut problem, which is solved by UV using an efficient solution~\cite{max_flow}.  Next, we introduce the algorithms for assigning weights to n-links and t-links.

\textbf{n-links} The weight $w(p_i, p_j)$ of the n-link between two adjacent patches $p_i$ and $p_j$ is given by~\cref{nlink_weight}.

\begin{equation} \label{nlink_weight}
\resizebox{6.5cm}{!}{$
\begin{aligned}
    &w(p_i, p_j) = \frac{e^{-\frac{(k_i - k_j)^T(k_i - k_j)}{\beta}}}{\sqrt{(i_x - j_x)^2 + (i_y - j_y)^2}}\\
    &\beta = \frac{2\sum_{p_i \in P}\sum_{p_j \in P}\mathbf{1}(p_i, p_j)(k_i - k_j)^T(k_i - k_j)}{C}
\end{aligned}
$}
\end{equation}

where $(i_x, i_y)$ and $(j_x, j_y)$ are the indices of the two patches under the scale of the feature map (\ie $28 \times 28$) output by the ViT. $\mathbf{1}(p_i, p_j)$ is an indicator function which returns 1 if the $p_i$ and $p_j$ are adjacent, otherwise 0. Finally, $C$ is the number of n-links in the graph.

\textbf{t-links} An UV\textquotesingle s graph has three kinds of patches: seed patch $p_f$, anti-seed patches set $B_f$, and the other patches $P-B_f\bigcup\{p_f\}$, whose weight assignment strategies for t-links are different. Suppose the Source and Target are defined as $S$ and $T$. After min-cut, the t-link from $S$ to $p_f$ should be kept while the t-link between $p_f$ and $T$ should be cut. To this end, a large weight $W$ is given to the t-link from $S$ to $p_f$ so that it is too large to be cut, and the weight between $p_f$ and $T$ is 0. Following early graph cut methods~\cite{grabcut, graph_cut}, $W$ can be decided by~\cref{K}.

\begin{equation} \label{K}
\resizebox{6cm}{!}{$
W = 1 + \mathop{\max}_{p_i \in P} \sum_{p_j \in P}\mathbf{1}(p_i, p_j)w(p_i, p_j)
$}
\end{equation}

Similarly, anti-seed patches must stay connected with $T$ after min-cut. Therefore, t-links from $\forall p_b \in B_f$ to $T$ must have a large value, which can also be given by~\cref{K}, while the weights of t-links from $S$ to $\forall p_b \in B_f$ are 0.

Weights of other t-links between $S$ or $T$ and $p_i \notin B_f\bigcup\{p_f\}$ are decided by~\cref{t_link_fore}.

\begin{equation} \label{t_link_fore}
\resizebox{7.5cm}{!}{$
\begin{aligned}
w(p_i, S) &= - \log \frac{e^{\sqrt{(k_i - k_f)^T(k_i - k_f)}}}{e^{\sqrt{(k_i - k_f)^T(k_i - k_f)}} + e^{\mathop{min}_{p_j \in B_f}\sqrt{(k_i - k_j)^T(k_i - k_j)}}}\\
w(p_i, T) &= - \log \frac{e^{\mathop{min}_{p_j \in B_f}\sqrt{(k_i - k_j)^T(k_i - k_j)}}}{e^{\sqrt{(k_i - k_f)^T(k_i - k_f)}} + e^{\mathop{min}_{p_j \in B_f}\sqrt{(k_i - k_j)^T(k_i - k_j)}}}
\end{aligned}
$}
\end{equation}
\cref{t_link_fore} explains that if a patch $p_i \notin B_f\bigcup \{p_f\}$ is more similar to anti-seed patches set $B_f$ than the seed patch $p_f$, $w(p_i, T)$ should exceed $w(p_i, S)$, making the t-link from $S$ to $p_i$ more likely to be cut, and vice versa.

With the graph initialized using the method introduced above, the UV executes a min-cut and returns a binary mask, where ``1" indicates patches remaining connected to the Source, \ie regions sharing features with the seed patch $p_f$. Therefore, the function of each UV is to return the region that is similar to its seed patch.

\subsection{UnionCut}
\label{sec:union_cut}
UnionCut is a strong classifier composed of 784 UVs as weak classifiers, with each UV using a unique patch from the image\textquotesingle s $28 \times 28$ patches as its seed patch. Considering the function of UV introduced above, there are two types of UVs in UnionCut: 1) background UV: if its seed patch is background, the UV returns a binary mask indicating background regions; and 2) foreground UV: if its seed patch is foreground, the UV outputs a binary mask of foreground regions. UnionCut\textquotesingle s output is determined by voting from the 784 UVs. Given the set of binary masks  $M=\{m_f|\forall p_f\in P\}$ returned by all UVs, a heat map $A$ is generated by aggregating all UVs\textquotesingle ~output using~\cref{aggregating}.
\begin{equation} \label{aggregating}
\resizebox{4cm}{!}{$
A(x, y) = \sum_{m_f \in M} m_f(x, y)
$}
\end{equation}
where $x\in[1, 28]$ and $y \in[1,28]$ indicate the 2D coordinate of a patch\textquotesingle s response on $A$.

For easier understanding, intuitively, the foreground usually occupies a smaller area than the background in the image~\cite{simeoni2021localizing}. Thus, we can assume that there are more background UVs than foreground UVs in UnionCut, causing background areas to have higher intensities on the aggregated heat map $A$. An example can be seen in~\cref{unioncut_framework}. To make it intuitive, the brightness of $A$ is inverted so that foreground regions have higher responses than the background, producing an inverted heat map $H$ using ~\cref{inverse}. By thresholding $H$, most objects in the image are segmented as the foreground union.

\begin{equation} \label{inverse}
\resizebox{7cm}{!}{$
H(x, y) = 255 - 255 \times \frac{A(x, y) - \mathop{\min}_{x, y}A(x, y)}{\mathop{\max}_{x, y}A(x, y) - \mathop{\min}_{x, y}A(x, y)}\\
$}
\end{equation}

Note that the assumption introduced above is for easier understanding of our approach. UnionCut is mathematically and statistically robust, with an explanation in Appendix~\ref{prove}. Moreover, UnionCut remains effective even when the foreground area is larger than the background, with examples in Appendix~\ref{large_foreground_example}. The mathematical explanation also supports this claim. Furthermore, compared to more straightforward methods detecting patches similar to a given patch, \eg matching with cosine similarity~\cite{simeoni2021localizing, FOUND}, UV is more sophisticated. Appendix~\ref{UBVvsCOS} provides a detailed analysis of UV\textquotesingle s irreplaceable advantages and necessity.

\subsubsection{Thresholding}
A threshold for the bipartition of the heat map is required to estimate the foreground union. UnionCut automatically determines a threshold for the heat map $H$ by clustering the response values on it with Mean-Shift~\cite{mean_shift}. Since the bright areas on the heat map represent the foreground, the resulting clusters are sorted by their centres in descending order, with the top half retained and identified as the union of the foreground in the image. Finally, a $28 \times 28$ binary mask $U_{cut}$ with ``1" as the foreground union is obtained.

\subsubsection{Corner Prior}
Similar to~\cite{wang2023cut}, a corner prior is used by UnionCut to check the reliability of the foreground union mask. Specifically, if $U_{cut}$ occupies four corners of the image, it must be inverted (\ie $U_{cut}\coloneqq1-U_{cut}$) because the foreground union is unlikely to occupy four corners simultaneously. The reliability of this corner prior is discussed in Appendix~\ref{reliability_corner}.

\subsection{UnionSeg}
While UnionCut effectively detects an image\textquotesingle s foreground union, running 784 UVs per image is computationally expensive. To improve efficiency, we introduce UnionSeg, a ViT-based model distilled from UnionCut as a surrogate model. Unlike UnionCut, UnionSeg is an end-to-end model that directly predicts the foreground union from an input image. To minimize learnable parameters, inspired by FOUND~\cite{FOUND}, UnionSeg uses a frozen ViT trained by DINO~\cite{dino} as its backbone, followed by a learnable $1 \times 1$ convolutional layer and a sigmoid function, compressing each ViT patch feature into a scalar confidence score. Patches with scores above 0.5 are classified as the foreground union. Appendix~\ref{UnionSeg_vs_FOUND} details the differences between UnionSeg and FOUND. Suppose the soft output of UnionSeg of an image is $U_{seg}^{s}$, with its thresholded hard output defined as $U_{seg}^{h}$. Given UnionCut\textquotesingle s output $U_{cut}$ of an image, the label $L$ of an image used to train UnionSeg is defined by~\cref{pl}.
\begin{equation}\label{pl}
\resizebox{5cm}{!}{$
L = \begin{cases}
U_{cut}, & \text{if } IoU(U_{seg}^{h}, U_{cut}) < 0.5, \\
U_{seg}^{h}, & \text{otherwise.}
\end{cases}
$}
\end{equation}
where $IoU(\cdot)$ returns the intersection over union between two binary masks. The motivation behind ~\cref{pl} is that UnionCut\textquotesingle s output may contain errors, so UnionSeg does not rely entirely on UnionCut\textquotesingle s output as labels. This enables UnionSeg to predict the foreground union more efficiently and accurately than UnionCut. The loss function used to train UnionSeg is shown in~\cref{loss}.

\begin{equation}\label{loss}
\resizebox{7cm}{!}{$
\begin{aligned}
    &Loss = L\log U_{seg}^{s} + (1-L) \log(1-U_{seg}^{s}) + \mathbf{1}(iter<100)\Lambda\\
    &\Lambda = U_{cut}\log U_{seg}^{s} + (1-U_{cut}) \log(1-U_{seg}^{s})
\end{aligned}
$}
\end{equation}
where $\mathbf{1}(iter<100)$ returns 1 if the training iterations are fewer than 100; otherwise 0. This ensures that at the start of training, when UnionSeg is not yet well trained, it is forced to use both UnionCut\textquotesingle s output $U_{cut}$ and $L$ as labels to aid learning.

\section{Experiments}
\label{sec:experiment}
Several experiments are conducted to evaluate the effectiveness of UnionCut/UnionSeg, comparing the performance of state-of-the-art UOD methods before and after integrating UnionCut or UnionSeg on several tasks: unsupervised single object discovery (\cref{subsec:uod}), unsupervised saliency detection (\cref{subsec:usd}), and self-supervised instance segmentation (\cref{subsec:ssis}). We then discuss the properties of UnionCut and UnionSeg in \cref{subsec:puu}, and evaluate the impact of different components of them in \cref{subsec:as}. Additionally, we provide more qualitative comparisons in Appendix~\ref{more_vis}.

\textbf{Baseline UOD methods} Our experiments use existing 
state-of-the-art UOD methods for single or multiple object discovery as baselines, \ie LOST~\cite{simeoni2021localizing}, TokenCut~\cite{wang2022tokencut}, FOUND~\cite{FOUND} and CutLER~\cite{wang2023cut}. To enhance their performance, we integrate them with UnionCut and UnionSeg by replacing their standard foreground priors with UnionCut/UnionSeg, with details provided in Appendix~\ref{implementation}.

\begin{table}[t]
    \centering
    \footnotesize
    \renewcommand\tabcolsep{3pt}
    \renewcommand\arraystretch{1.1}
    \begin{tabular}{lccc}
        \toprule
        Method & VOC07 & VOC12 & COCO20K\\
        \hline
        \multicolumn{4}{c}{\textbf{- No learning -}}\\
        Selective Search~\cite{ss} & 18.8 & 20.9 & 16.0\\
        EdgeBoxes~\cite{edgeboxes} & 31.1 & 31.6 & 28.8\\
        Kim~\etal~\cite{kim} & 43.9 & 46.4 & 35.1\\
        Zhang~\etal~\cite{zhang} & 46.2 & 50.5 & 34.8\\
        DDT+~\cite{DDT} & 50.2 & 53.1 & 38.2\\
        rOSD~\cite{rOSD} & 54.5 & 55.3 & 48.5\\
        LOD~\cite{LOD} & 53.6 & 55.1 & 48.5\\
        DINO-seg~\cite{dino,simeoni2021localizing}(ViT-S/16~\cite{dino}) & 45.8 & 46.2 & 42.0\\
        DSS~\cite{DSS}(ViT-S/16~\cite{dino}) & 62.7 & 66.4 & 52.2\\
        \hdashline
        LOST~\cite{simeoni2021localizing}(ViT-S/16~\cite{dino}) & 61.9 & 64.0 & 50.7\\
        \textbf{LOST(ViT-S/16~\cite{dino})+UnionCut} & 62.4\textbf{(0.5$\uparrow$)} & 65.3\textbf{(1.3$\uparrow$)} & 51.9\textbf{(1.2$\uparrow$)}\\
        \textbf{LOST(ViT-S/16~\cite{dino})+UnionSeg} & 63.2\textbf{(1.3$\uparrow$)} & 67.1\textbf{(3.1$\uparrow$)} & 54.9\textbf{(4.2$\uparrow$)}\\
        \hdashline
        TokenCut~\cite{wang2022tokencut}(ViT-S/16~\cite{dino}) & 68.8 & 72.1 & 58.8\\
        \textbf{TokenCut(ViT-S/16~\cite{dino})+UnionCut} & 69.2\textbf{(0.4$\uparrow$)} & 72.3\textbf{(0.2$\uparrow$)} & 62.1\textbf{(3.3$\uparrow$)}\\
        \textbf{TokenCut(ViT-S/16~\cite{dino})+UnionSeg} & \textcolor{red}{69.7}\textbf{(0.9$\uparrow$)} & \textcolor{red}{72.7}\textbf{(0.6$\uparrow$)} & \textcolor{red}{62.6}\textbf{(3.8$\uparrow$)}\\
        \hline
        \multicolumn{4}{c}{\textbf{- With learning -}}\\
        FreeSOLO~\cite{freesolo,wang2023cut} & 44.0 & 49.7 & 35.2\\
        LOD+CAD~\cite{simeoni2021localizing, wang2022tokencut} & 56.3 & 61.6 & 52.7\\
        rOSD+CAD~\cite{simeoni2021localizing, wang2022tokencut} & 58.3 & 62.3 & 53.0\\
        LOST+CAD~\cite{simeoni2021localizing}(ViT-S/16~\cite{dino}) & 65.7 & 70.4 & 57.5\\
        SelfMask~\cite{selfmask} & 72.3 & 75.3 & 62.7\\
        \hdashline
        FOUND~\cite{FOUND}(ViT-S/16~\cite{dino}) & 72.5 & \textcolor{red}{76.1} & 62.9\\
        FOUND(ViT-S/16~\cite{dino})$\ddagger$ & 71.4 & 75.7 & 62.9\\
        \textbf{FOUND(ViT-S/16~\cite{dino})+UnionCut} & 72.6\textbf{(1.2$\uparrow$)} & \textcolor{red}{76.1}\textbf{(0.4$\uparrow$)} & 63.0\textbf{(0.1$\uparrow$)}\\
        \textbf{FOUND(ViT-S/16~\cite{dino})+UnionSeg} & 72.3\textbf{(0.9$\uparrow$)} & 75.9\textbf{(0.2$\uparrow$)} & 63.1\textbf{(0.2$\uparrow$)}\\
        \hdashline
        CutLER~\cite{wang2023cut}(ViT-B/8~\cite{dino})$\dagger$ & 73.3 & 69.5 & 70.7\\
        \textbf{CutLER(ViT-B/8~\cite{dino})+UnionSeg} & \textcolor{red}{73.8}\textbf{(0.5$\uparrow$)} & 71.2\textbf{(1.7$\uparrow$)} & \textcolor{red}{72.4}\textbf{(1.7$\uparrow$)}\\
        \bottomrule
    \end{tabular}
    \caption{\textbf{Performances of UOD methods on single object discovery with \textcolor{red}{red} indicates the highest score.} CAD: a class-agnostic detector trained with the output of no-learning UOD methods as the ground truth. $\dagger$: results of official implementations and checkpoints by our measurement. $\ddagger$: results of our replication with official implementation.}\label{corloc}
\end{table}

\begin{table*}[t]
    \centering
    \scriptsize
    \renewcommand\tabcolsep{5pt}
    \renewcommand\arraystretch{1}
    \begin{tabular}{lccccccccc}
        \toprule
        & \multicolumn{3}{c}{DUT-OMRON} & \multicolumn{3}{c}{DUTS-TE} & \multicolumn{3}{c}{ECSSD} \\
        \cmidrule(lr){2-4} \cmidrule(lr){5-7} \cmidrule(lr){8-10}
        Method & Acc. & IoU & maxF\textsubscript{$\beta$} & Acc. & IoU & maxF\textsubscript{$\beta$} & Acc. & IoU & maxF\textsubscript{$\beta$} \\
        \hline
        \multicolumn{10}{c}{\textbf{- No learning -}}\\
        HS~\cite{hs} & 84.3 & 43.3 & 56.1 & 82.6 & 36.9 & 50.4 & 84.7 & 50.8 & 67.3 \\
        wCtr~\cite{wctr} & 83.8 & 41.6 & 54.1 & 83.5 & 39.2 & 52.2 & 86.2 & 51.7 & 68.4 \\
        WSC~\cite{WSC} & 86.5 & 38.7 & 52.3 & 86.2 & 38.4 & 52.8 & 85.2 & 49.8 & 68.3 \\
        DeepUSPS~\cite{deepusps} & 77.9 & 30.5 & 41.4 & 77.3 & 30.5 & 42.5 & 79.5 & 44.0 & 58.4 \\
        BigBiGAN~\cite{BigBiGAN} & 85.6 & 45.3 & 54.9 & 87.8 & 49.8 & 60.8 & 89.9 & 67.2 & 78.2 \\
        E-BigBiGAN~\cite{BigBiGAN} & 86.0 & 46.4 & 56.3 & 88.2 & 51.1 & 62.4 & 90.6 & 68.4 & 79.7 \\
        LOST~\cite{simeoni2021localizing} & 79.7 & 41.0 & 47.3 & 87.1 & 51.8 & 61.1 & 89.5 & 65.4 & 75.8 \\
        DSS~\cite{DSS} & - & 56.7 & - & - & 51.4 & - & - & 73.3 & - \\
        MOST~\cite{most} & 87 & 47.5 & 57.0 & 89.7 & 53.8 & 66.6 & 89.0 & 63.1 & 79.1 \\
        \hdashline
        TokenCut~\cite{wang2022tokencut,FOUND}(ViT-S/16~\cite{dino}) & 89.7 & 61.8 & 69.7 & 91.4 & 62.4 & 75.5 & 93.4 & 77.2 & 87.4 \\
        \textbf{TokenCut(ViT-S/16~\cite{dino})+UnionCut} & 89.6 & \textcolor{red}{62.4}\textbf{(0.6$\uparrow$)} & 71.0\textbf{(1.3$\uparrow$)} & 91.6\textbf{(0.2$\uparrow$)} & 63.2\textbf{(0.8$\uparrow$)} & 80.2\textbf{(4.7$\uparrow$)} & 93.8\textbf{(0.4$\uparrow$)} & \textcolor{red}{77.3}\textbf{(0.1$\uparrow$)} & \textcolor{red}{91.0}\textbf{(3.6$\uparrow$)} \\
        \textbf{TokenCut(ViT-S/16~\cite{dino})+UnionSeg} & \textcolor{red}{90.6}\textbf{(0.9$\uparrow$)} & 62.1\textbf{(0.3$\uparrow$)} & \textcolor{red}{71.2}\textbf{(1.5$\uparrow$)} & \textcolor{red}{92.3}\textbf{(0.9$\uparrow$)} & \textcolor{red}{63.4}\textbf{(1.0$\uparrow$)} & \textcolor{red}{80.6}\textbf{(5.1$\uparrow$)} & \textcolor{red}{94.1}\textbf{(0.7$\uparrow$)} & 77.2 & \textcolor{red}{91.0}\textbf{(3.6$\uparrow$)} \\
        \hline
        \multicolumn{10}{c}{\textbf{- With learning -}}\\
        LOST~\cite{simeoni2021localizing} & 81.8 & 48.9 & 57.8 & 88.7 & 57.2 & 69.7 & 91.6 & 72.3 & 83.7 \\
        SelfMask~\cite{selfmask} & 91.9 & 65.5 & - & 93.3 & 66.0 & - & \textcolor{red}{95.5} & \textcolor{red}{81.8} & - \\
        UnSeGArmaNet~\cite{UnSeGArmaNet} & - & - & - &  & 60.3 & - & - & 77.8 & - \\
        \hdashline
        CutLER~\cite{wang2023cut}(ViT-B/8~\cite{dino})$\dagger$ & 74.3 & 43.8 & 48.1 & 78.1 & 46.5 & 54.2 & 87.2 & 69.3 & 76.3 \\
        \textbf{CutLER(ViT-B/8~\cite{dino})+UnionSeg} & 87.7\textbf{(13.4$\uparrow$)} & 57.8\textbf{(14.0$\uparrow$)} & 64.1\textbf{(16.0$\uparrow$)} & 90.1\textbf{(12.0$\uparrow$)} & 61.4\textbf{(14.9$\uparrow$)} & 72.3\textbf{(18.1$\uparrow$)} & 94.0\textbf{(6.8$\uparrow$)} & 79.6\textbf{(10.3$\uparrow$)} & 87.8\textbf{(11.5$\uparrow$)} \\
        \hdashline
        FOUND~\cite{FOUND}(ViT-S/16~\cite{dino}) & 92.2 & 61.3 & 70.8 & 94.2 & 66.3 & 76.3 & 95.1 & 81.3 & \textcolor{red}{93.5} \\
        FOUND(ViT-S/16~\cite{dino})$\ddagger$ & 91.8 & 60.8 & 69.4 & 94.0 & 66.0 & 75.4 & 95.0 & 81.0 & 93.0 \\
        \textbf{FOUND(ViT-S/16~\cite{dino})+UnionCut} & 92.4\textbf{(0.6$\uparrow$)} & \textcolor{red}{61.8}\textbf{(1.0$\uparrow$)} & 71.2\textbf{(1.8$\uparrow$)} & \textcolor{red}{94.4}\textbf{(0.4$\uparrow$)} & \textcolor{red}{66.7}\textbf{(0.7$\uparrow$)} & 76.7\textbf{(1.3$\uparrow$)} & 94.9 & 80.6 & 93.3\textbf{(0.3$\uparrow$)} \\
        \textbf{FOUND(ViT-S/16~\cite{dino})+UnionSeg} & \textcolor{red}{92.5}\textbf{(0.7$\uparrow$)} & 61.7\textbf{(0.9$\uparrow$)} & \textcolor{red}{71.7}\textbf{(2.3$\uparrow$)} & \textcolor{red}{94.4}\textbf{(0.4$\uparrow$)} & 66.6\textbf{(0.6$\uparrow$)} & \textcolor{red}{77.1}\textbf{(1.7$\uparrow$)} & 94.7 & 80.2 & 93.4\textbf{(0.4$\uparrow$)} \\
        \bottomrule
    \end{tabular}
    \caption{\textbf{Performances of UOD methods on saliency detection with \textcolor{red}{red} indicates the highest score.} $\dagger$: results of official codes and checkpoints by our measurement. $\ddagger$: results of our replication with official implementation.}\label{saliency}
\end{table*}

\textbf{Implementation Details} We use a ViT-S/8~\cite{vit} pretrained by DINO~\cite{dino} as the image feature extractor for UnionCut and UnionSeg. Following \cite{selfmask, FOUND}, we distill UnionSeg from UnionCut using DUTS-TR\cite{duts}, a saliency detection dataset with 10,553 images. Training is done with a batch size of 50, an initial learning rate of 0.05 that decays to 95\% every 50 iterations, and AdamW~\cite{adamW} for parameter updates over 600 iterations. Implementation details for all tested methods, \ie LOST+UnionCut/UnionSeg, TokenCut+UnionCut/UnionSeg, FOUND+UnionCut/UnionSeg, and CutLER+UnionSeg, are provided in Appendix~\ref{implementation}.

\subsection{Unsupervised Single Object Discovery}
\label{subsec:uod}
In this experiment, we combine baseline UOD methods with UnionCut and UnionSeg to see if their performance improves in unsupervised single object discovery.

\textbf{Benchmark} Following~\cite{simeoni2021localizing, wang2022tokencut, FOUND}, the \textit{trainval} sets of PASCAL VOC07 (5011 images), VOC12 (11540 images)~\cite{voc} and COCO20K~\cite{rOSD, coco} (19817 randomly sampled images from COCO2014~\cite{coco}) are selected as the benchmark for evaluation.

\textbf{Metrics} Similar to~\cite{LOD, rOSD,simeoni2021localizing, wang2022tokencut, FOUND}, we use CorLoc (Correct Localization) to evaluate UOD algorithms\textquotesingle ~performance on single object discovery. CorLoc scores an image as 1 if the intersection over union (IoU) between any ground truth bounding box and the algorithm\textquotesingle s predicted bounding box exceeds 0.5, otherwise it scores 0. In our experiments, since CutLER can detect multiple objects, we select the bounding box with the highest confidence score as its single object prediction for the image.

\cref{corloc} shows that after integrating UnionCut or UnionSeg, our selected baseline UOD methods improve across all three benchmarks, achieving state-of-the-art results, indicating that UnionCut/UnionSeg as a robust foreground prior enhances UOD algorithms\textquotesingle ~performance in single object discovery. In Appendix~\ref{most}, we also evaluate these UOD methods from the perspective of the performance upper boundary proposed by Rambhatla~\etal~\cite{most}.

\subsection{Unsupervised Saliency Detection}
\label{subsec:usd}
We compare the performance of UOD methods before and after integrating UnionCut/UnionSeg on saliency detection, which involves segmenting salient objects in images. Since CutLER predicts multiple object masks per image, we use the union of these masks as its saliency detection result.

\textbf{Benchmark} Similar to~\cite{wang2022tokencut,FOUND,DSS,BigBiGAN,simeoni2021localizing}, ECSSD~\cite{ecssd} (1000 real-world images), DUTS-TE~\cite{duts} (5019 images sampled from ImageNet test), and DUT-OMRON~\cite{omron} (5168 natural images) are chosen as the benchmark.

\textbf{Metrics} Following~\cite{simeoni2021localizing, wang2022tokencut, FOUND}, we report results with three metrics: Accuracy, IoU, and maxF\textsubscript{$\beta$}. Accuracy measures the percentage of correctly classified pixels compared to ground truth. IoU is the ratio of the intersection between the UOD output and ground truth to their union. maxF\textsubscript{$\beta$} is calculated as $\frac{(1 + \beta^2)Precision \times Recall}{\beta^2Precision + Recall}$ where Precision is the percentage of predicted pixels matching ground truth, and Recall is the proportion of ground truth pixels covered by the prediction. We set $\beta$ to 0.3, as in~\cite{simeoni2021localizing, wang2022tokencut, FOUND}.

\cref{saliency} illustrates that the selected baseline methods improve when integrated with UnionCut/UnionSeg. TokenCut, with UnionSeg, performs best among no-learning approaches. And the performance of FOUND, the previous state-of-the-art method, is also increased by UnionCut and UnionSeg on most benchmarks and metrics.

\begin{table*}[htbp]
    \centering
    \scriptsize
    \renewcommand\tabcolsep{2pt}
    \renewcommand\arraystretch{1}
    \begin{tabular}{lccccccccccccccc}
        \toprule
        & & & & \multicolumn{6}{c}{COCO20K} & \multicolumn{6}{c}{COCO val2017} \\
        \cmidrule(lr){5-10} \cmidrule(lr){11-16}
        Methods & Pretrain & Detector & Init. & AP\textsubscript{50}\textsuperscript{box} & AP\textsubscript{75}\textsuperscript{box} & AP\textsuperscript{box} & AP\textsubscript{50}\textsuperscript{mask} & AP\textsubscript{75}\textsuperscript{mask} & AP\textsuperscript{mask} & AP\textsubscript{50}\textsuperscript{box} & AP\textsubscript{75}\textsuperscript{box} & AP\textsuperscript{box} & AP\textsubscript{50}\textsuperscript{mask} & AP\textsubscript{75}\textsuperscript{mask} & AP\textsuperscript{mask}\\
        \hline
        LOST~\cite{simeoni2021localizing} & IN+COCO & FRCNN~\cite{frcnn} & DINO & - & - & - & 2.4 & 1.0 & 1.1 & - & - & - & - & - & -\\
        MaskDistill~\cite{MaskDistill} & IN+COCO & MRCNN~\cite{mrcnn} & MoCo~\cite{moco} & - & - & - & 6.8 & 2.1 & 2.9 & - & - & - & - & - & -\\
        FreeSOLO~\cite{freesolo, wang2023cut} & IN+COCO & SOLOv2~\cite{solov2} & DenseCL~\cite{denseCL} & 9.7 & 3.2 & 4.1 & 9.7 & 3.4 & 4.3 & 9.6 & 3.1 & 4.2 & 9.4 & 3.3 & 4.3\\
        DETReg~\cite{detreg} & IN & DDETR~\cite{ddetr} & SwAV~\cite{swav} & - & - & - & - & - & - & 3.1 & 0.6 & 1.0 & 8.8 & 1.9 & 3.3\\
        DINO~\cite{dino} & IN & - & DINO & 1.7 & 0.1 & 0.3 & - & - & - & - & - & - & - & - & -\\
        ProMerge~\cite{ProMerge} & IN & - & DINO & - & - & - & - & - & 3.0 & - & - & - & - & - & 2.4\\
        ProMerge+~\cite{ProMerge} & IN & Cascade~\cite{cascade} & DINO & - & - & - & - & - & 8.9 & - & - & - & - & - & 9.0\\
        \hdashline
        CutLER~\cite{wang2023cut} & IN & Cascade~\cite{cascade} & DINO & 22.4 & \textbf{12.5} & 11.9 & 19.6 & \textbf{10.0} & 9.2 & 21.9 & 11.8 & 12.3 & 18.9 & \textbf{9.7} & 9.2\\
        \textbf{CutLER+UnionSeg} & IN & Cascade~\cite{cascade} & DINO & \textbf{24.1} & 12.2 & \textbf{12.9} & \textbf{20.3} & 9.1 & \textbf{10.0} & \textbf{23.7} & \textbf{12.2} & \textbf{12.9} & \textbf{19.7} & 9.0 & \textbf{9.9}\\
        \bottomrule
    \end{tabular}
    \caption{\textbf{Performances of self-supervised instance segmentation.} The dataset (Pretrain) used to train each model (Detector) and each model\textquotesingle s initialization (Init) are also listed. IN denotes ImageNet.}
    \label{instance_all}
\end{table*}



\subsection{Self-supervised Instance Segmentation}
\label{subsec:ssis}
Some UOD methods target self-supervised instance segmentation, where models are trained using pseudo-annotations given by UOD algorithms, such as MaskCut for CutLER and FreeMask for FreeSOLO~\cite{freesolo}. In the experiment, we explore whether UnionSeg improves CutLER\textquotesingle s performance on instance segmentation. Similar to~\cite{wang2023cut}, COCO20k~\cite{rOSD, coco} and COCO~\cite{coco}(5000 images from COCO2017\textquotesingle s validation set) are used for evaluation, with average precision (AP) as the metric. \cref{instance_all} illustrates that by using UnionSeg to improve the pseudo-labels given by MaskCut, CutLER performs better on most metrics.

Since UnionCut with MaskCut takes much time to generate the foreground union and UOD results of all images in ImageNet (7 weeks in total, with 4 weeks for UnionCut and 3 weeks for MaskCut with 18 subprocesses running in parallel limited by our available computational resources), we only consider UnionSeg in this section. However, the effectiveness of UnionCut on self-supervised instance segmentation is also of interest. Therefore, we conduct another experiment that trains CutLER on a much smaller dataset (the subset of VOC12 containing 2913 images with pixel-level annotations). Details can be found in Appendix~\ref{unioncut_instance}.



\subsection{Properties of UnionCut and UnionSeg}
\label{subsec:puu}
We explore the properties of UnionCut and UnionSeg. First, their accuracy in foreground union detection is evaluated.

\textbf{Baseline} Li~\etal~\cite{ProMerge} proposes ProMerge, which contains a foreground union detection method. FOUND~\cite{FOUND} is a UOD method; however, considering that it includes an algorithm for detecting background regions in an image, we extract this algorithm separately as a baseline and treat the inversion of its output as the foreground union detection result. For convenience, we still refer to their foreground union detection algorithms as ProMerge and FOUND.

\textbf{Benchmark} The subset of VOC12 with 2913 images of pixel-level annotations is used as the benchmark, and the ground truth mask of each instance is combined as the ground truth of the foreground union in an image.

\textbf{Metrics} IoU, precision, and recall are used to measure the performance. Besides, since representative datasets~\cite{voc, coco} label objects of a given class list, there may be cases where some objects in the image are not labelled and thus are not considered part of the foreground union by the ground truth (Examples can be seen in Appendix~\ref{sec:voc_example}). Therefore, directly using these metrics and dataset-provided labels to measure the performance of foreground union detection is inaccurate. To address this, we propose a fuzzy evaluation metric similar to CorLoc, named CorUnion. Specifically, we compare the foreground union output by UnionCut or UnionSeg with the ground truth for a given image and compute a specific metric (\eg IoU). If this metric exceeds a certain threshold, we consider the foreground union detection on that image successful. The success rate across the dataset is then used as the CorUnion value. This approach also allows us to study the properties of UnionCut and UnionSeg by using different metrics or adjusting the threshold for successful detection.

\begin{figure}[htbp]
    \centering
    \includegraphics[width=7cm]{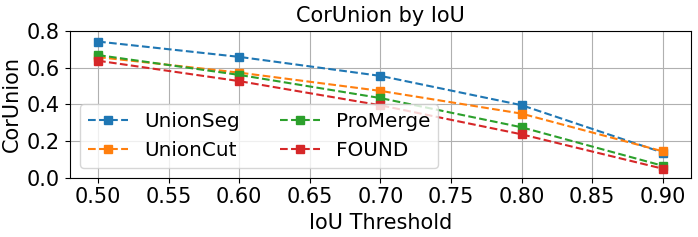}
    \caption{\textbf{Foreground union detection measured by CorUnion.}}
    \label{accuracy}
\end{figure}

\textbf{Foreground Union Accuracy} \cref{rebuttal} shows the performance of UnionCut and UnionSeg, measured by precision, recall, and IoU. We also report CorUnion using IoU as the metric to judge an image\textquotesingle s foreground union detection success, with increasing IoU threshold from 0.5 to 0.9. As shown in \cref{accuracy}, both UnionCut and UnionSeg are more accurate than the baselines at foreground union detection.

\begin{table}[htbp]
    \centering
    \scriptsize
    \renewcommand\tabcolsep{14pt}
    \renewcommand\arraystretch{1}
    \begin{tabular}{lccc}
        \toprule
        Method & Precision & Recall & IoU\\
        \hline
        FOUND & 69.2 & 81.1 & 57.9\\
        ProMerge & 73.4 & 78.4 & 59.9\\
        \hdashline
        UnionCut & 70.7 & \textcolor{red}{84.6} & 60.9\\
        UnionSeg & \textcolor{red}{80.9} & 78.6 & \textcolor{red}{65.7}\\
        \bottomrule
    \end{tabular}
    \caption{\textbf{Average performance on foreground union detection.}}
    \label{rebuttal}
\end{table}


\begin{figure}[htbp]
    \centering
    \begin{subfigure}[b]{\linewidth}
        \centering
        \includegraphics[width=\linewidth]{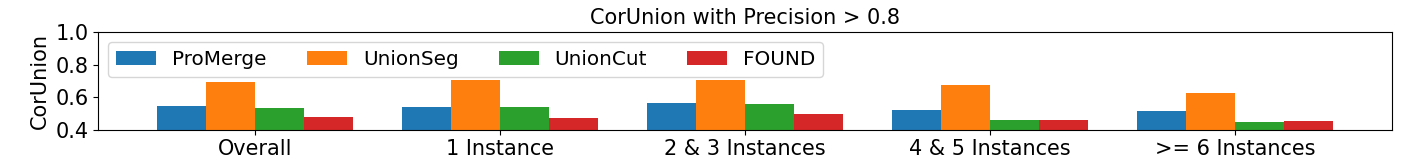}
        \caption{}
        \label{pr-a}
    \end{subfigure}
    \hfill
    \begin{subfigure}[b]{\linewidth}
        \centering
        \includegraphics[width=\linewidth]{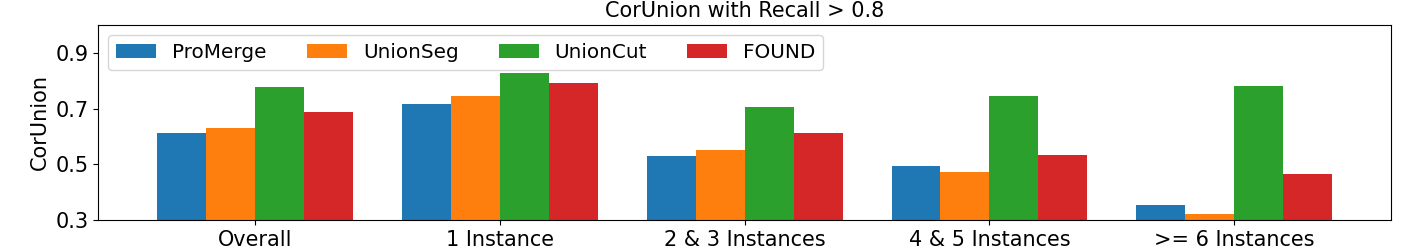}
        \caption{}
        \label{pr-b}
    \end{subfigure}
    \caption{\textbf{Performance of foreground union detection on images with varying instance numbers, measured by CorUnion with precision$>$0.8 (a) and recall$>$0.8 (b) as success criteria.}}
    \label{pr}
\end{figure}

\textbf{Instances Count} In addition, we assessed UnionCut and UnionSeg\textquotesingle s performance across varying instance counts in VOC12, categorized into groups based on the number of instances per image. As shown in \cref{pr-a}, UnionSeg, a training-based method, outperforms training-free approaches in precision, supporting that self-training corrects initial errors made by training-free methods. This finding aligns with previous research~\cite{wang2022tokencut,FOUND,simeoni2021localizing}. As instance numbers increase, UnionSeg maintains its precision, while other methods decline. Furthermore, \cref{pr-b} shows that UnionCut surpasses other methods in recall of the foreground union, remaining stable as the number of instances rises, unlike the decrease seen with other methods. This demonstrates the functional distinction between UnionCut and other training-free methods: as the number of instances per image increases, UnionCut consistently predicts the foreground union covering most instances, though precision inevitably decreases. This consistency highlights UnionCut\textquotesingle s effectiveness as a robust prior for UOD algorithms to cease exploration optimally.

Overall, UnionSeg outperforms UnionCut and other baseline methods (\cref{accuracy}) with higher precision (\cref{pr-a}), making it better suited for identifying if a detected area is foreground. UnionCut, with its higher recall (\cref{pr-b}), is better for determining if the discovery is complete.

\subsection{Ablation Study}
\label{subsec:as}
UnionCut consists of two elements: aggregated UVs and corner prior, while UnionSeg builds on UnionCut with distillation. To evaluate each element\textquotesingle s contribution, TokenCut~\cite{wang2022tokencut} is used as the baseline, with each element integrated for single object discovery. Benchmarks and metrics are inherited from~\cref{subsec:uod}. As shown in \cref{ablation}, each element contributes to the performance increase of UOD, demonstrating their effectiveness. 

\begin{table}[htbp]
    \centering
    \footnotesize
    \renewcommand\tabcolsep{3pt}
    \renewcommand\arraystretch{1}
    \begin{tabular}{lcccccc}
        \toprule
        Method & aU & cp & d & VOC07 & VOC12 & COCO20K\\
        \hline
        TokenCut~\cite{wang2022tokencut} & & & & 68.8 & 72.1 & 58.8\\
        TokenCut+aU & \Checkmark & & & 69.0 & 72.3 & 62.0\\
        TokenCut+UnionCut & \Checkmark & \Checkmark & & 69.2 & 72.3 & 62.1\\
        TokenCut+UnionSeg & \Checkmark & \Checkmark & \Checkmark & \textbf{69.7} & \textbf{72.7} & \textbf{62.6}\\
        \bottomrule
    \end{tabular}
    \caption{\textbf{Contribution of each component of UnionCut and UnionSeg.} aU, cp, and d denote aggregated UVs, corner prior and distillation, respectively.}\label{ablation}
\end{table}

We further evaluate the inference speed of UnionCut and UnionSeg. We include FOUND and ProMerge in \cref{subsec:puu} as the baseline. Our implementation is based on Python 3.11.9, Pytorch 2.3.1 and CUDA 12.1. As shown in \cref{fps}, UnionSeg significantly increases the efficiency of foreground union detection compared with UnionCut.

\begin{table}[htbp]
    \centering
    \footnotesize
    \renewcommand\tabcolsep{5pt}
    \begin{tabular}{lcccccc}
        \toprule
        Method & CPU & GPU & FPS\\
        \hline
        FOUND & Intel 14700KF & NVIDIA RTX4070Ti Super & 2\\
        ProMerge & Intel 14700KF & NVIDIA RTX4070Ti Super & 3\\
        \hdashline
        UnionCut & Intel 14700KF & NVIDIA RTX4070Ti Super & 0.1\\
        UnionSeg & Intel 14700KF & NVIDIA RTX4070Ti Super & 125\\
        \bottomrule
    \end{tabular}
    \caption{\textbf{Inference speed of UnionSeg and UnionCut}. FPS (frames per second) is evaluated throughout the whole pipeline from the image input to the output of the foreground union.}\label{fps}
\end{table}
\section{Conclusion}
\label{sec:conclusion}
This paper introduces UnionCut, an unsupervised ensemble method to robustly predict foreground unions in images. Additionally, we propose UnionSeg distilled from UnionCut to detect foreground unions more accurately and efficiently. Both of them help UOD algorithms determine if discovered areas are in the foreground and when to stop discovery on an image, enhancing UOD\textquotesingle s accuracy. While both are effective, each has unique advantages, and the choice between them depends on the specific UOD requirements for precision, recall, and efficiency. Our experiments show that integrating UnionCut or UnionSeg improves the performance of state-of-the-art UOD methods in single object discovery, saliency detection, and instance segmentation, demonstrating their effectiveness as robust foreground priors for unsupervised object discovery.
{
    \small
    \bibliographystyle{ieeenat_fullname}
    \bibliography{main}
}

\clearpage
\setcounter{page}{1}
\maketitlesupplementary

\section{Mathematical and Statistical Explanation for UnionCut}
\label{prove}
In this section, we mathematically discuss the effectiveness of UnionCut and explain why it works. Suppose $P=\{p_1, p_2, ...,p_{784}\}$ is the set of patches of an image, and $F$ and $B$ are subsets of $P$ representing the set of foreground patches and background patches given by the ground truth, respectively \ie $F\bigcup B = P$ and $F\bigcap B = \phi$. As for a Unit Voter (UV), its seed patch is denoted as $s_i \in P$, and the set of patches returned by the UV indicating the region similar to the seed patch $s_i$ is defined as $\hat{F}$. With the definitions above, after the execution of a UV with $s_i$ as the seed patch, for any background patch $\forall p_j \in B$, the probability of it being returned by the UV, \ie $p_j \in \hat{F}$, can be calculated by~\cref{pb}.

\begin{equation} \label{pb}
\resizebox{8.5cm}{!}{$
\begin{aligned}
    &P(p_j \in \hat{F}|p_j \in B)\\
    =&P(p_j \in \hat{F}, s_i \in F|p_j \in B) + P(p_j \in \hat{F}, s_i \in B|p_j \in B)\\
    =&P(s_i \in F) \cdot P(p_j \in \hat{F}|s_i \in F, p_j \in B) + P(s_i \in B) \cdot P(p_j \in \hat{F}|s_i \in B, p_j \in B)
\end{aligned}
$}
\end{equation}

Similarly, for a foreground patch $\forall p_j \in F$, its probability of being returned by the UV is given by~\cref{pf}.

\begin{equation} \label{pf}
\resizebox{8.5cm}{!}{$
\begin{aligned}
    &P(p_j \in \hat{F}|p_j \in F)\\
    =&P(p_j \in \hat{F}, s_i \in F|p_j \in F) + P(p_j \in \hat{F}, s_i \in B|p_j \in F)\\
    =&P(s_i \in F) \cdot P(p_j \in \hat{F}|s_i \in F, p_j \in F) + P(s_i \in B) \cdot P(p_j \in \hat{F}|s_i \in B, p_j \in F)
\end{aligned}
$}
\end{equation}

UnionCut is required to aggregate outputs of 784 UVs to generate a heat map indicating the background area in the image, with each patch on the image being selected as the seed by one of the 784 UVs. Since there is no ground truth and prior semantic knowledge of each patch available for UnionCut and each UV is conducted on its own graph with different seed patches and anti-seed patches, each execution of a UV is independent. Therefore, the mathematical expectation of the value corresponding to a foreground patch $P_j \in F$ or background patch $P_j \in B$ on the aggregated heat map $A$ given by~\cref{aggregating} in the paper can be calculated by~\cref{EF} or~\cref{EB}.

\begin{equation} \label{EF}
\resizebox{5cm}{!}{$
E(p_j \in \hat{F}|p_j \in F) = 784\cdot P(p_j \in \hat{F}|p_j \in F)
$}
\end{equation}

\begin{equation} \label{EB}
\resizebox{5cm}{!}{$
E(p_j \in \hat{F}|p_j \in B) = 784\cdot P(p_j \in \hat{F}|p_j \in B)
$}
\end{equation}

As introduced in~\cref{sec:union_cut}, UnionCut is expected to work on an image if background regions show higher responses than the foreground on the heat map $A$ generated by aggregating all UVs outputs. To achieve this goal, with~\cref{pb},~\cref{pf},~\cref{EF}, and~\cref{EB}, Inequality~\ref{ineq} should hold, where $P(s_i \in B)$ is of interest since it represents the prior probability that the selected seed $s_i \in P$ belongs to the background, which can be approximated by the proportion of background patches in the entire image.

\begin{figure*}[!ht]
\begin{equation} \label{ineq}
\resizebox{18cm}{!}{$
\begin{aligned}
&E(p_j \in \hat{F}|p_j \in B) > E(p_j \in \hat{F}|p_j \in F)\\
\Leftrightarrow &784\cdot P(p_j \in \hat{F}|p_j \in B) > 784\cdot P(p_j \in \hat{F}|p_j \in F)\\
\Leftrightarrow &P(p_j \in \hat{F}|p_j \in B) > P(p_j \in \hat{F}|p_j \in F)\\
\Leftrightarrow &P(s_i \in F) \cdot P(p_j \in \hat{F}|s_i \in F, p_j \in B) + P(s_i \in B) \cdot P(p_j \in \hat{F}|s_i \in B, p_j \in B) > P(s_i \in F) \cdot P(p_j \in \hat{F}|s_i \in F, p_j \in F) + P(s_i \in B) \cdot P(p_j \in \hat{F}|s_i \in B, p_j \in F)\\
\Leftrightarrow &P(s_i \in F) \cdot [P(p_j \in \hat{F}|s_i \in F, p_j \in B) - P(p_j \in \hat{F}|s_i \in F, p_j \in F)] > P(s_i \in B) \cdot [P(p_j \in \hat{F}|s_i \in B, p_j \in F) - P(p_j \in \hat{F}|s_i \in B, p_j \in B)]\\
\Leftrightarrow &[1 - P(s_i \in B)] \cdot [P(p_j \in \hat{F}|s_i \in F, p_j \in B) - P(p_j \in \hat{F}|s_i \in F, p_j \in F)] > P(s_i \in B) \cdot [P(p_j \in \hat{F}|s_i \in B, p_j \in F) - P(p_j \in \hat{F}|s_i \in B, p_j \in B)]\\
\Leftrightarrow &P(p_j \in \hat{F}|s_i \in F, p_j \in B) - P(p_j \in \hat{F}|s_i \in F, p_j \in F) > P(s_i \in B) \cdot [P(p_j \in \hat{F}|s_i \in B, p_j \in F) - P(p_j \in \hat{F}|s_i \in B, p_j \in B) + P(p_j \in \hat{F}|s_i \in F, p_j \in B) - P(p_j \in \hat{F}|s_i \in F, p_j \in F)]\\
\end{aligned}
$}
\end{equation}
\end{figure*}


\begin{table*}[t]
    \centering
    \footnotesize
    \begin{tabular}{lcccc}
        \toprule
        Dataset & $\hat{P}(p_j \in \hat{F}|s_i \in F, p_j \in B)$ & $\hat{P}(p_j \in \hat{F}|s_i \in B, p_j \in B)$ & $\hat{P}(p_j \in \hat{F}|s_i \in F, p_j \in F)$ & $\hat{P}(p_j \in \hat{F}|s_i \in B, p_j \in F)$\\
        \hline
        COCO2014 train~\cite{coco} & 0.1215 & 0.5496 & 0.1981 & 0.0563\\
        COCO20K~\cite{coco,simeoni2021localizing} & 0.1203 & 0.5499 & 0.1987 & 0.0572\\
        ECSSD~\cite{ecssd} & 0.0308 & 0.7270 & 0.1927 & 0.0198\\
        DUTS-TR~\cite{duts} & 0.0495 & 0.7628 & 0.2098 & 0.0264\\
        DUTS-TE~\cite{duts} & 0.0549 & 0.6621 & 0.2099 & 0.0294\\
        \bottomrule
    \end{tabular}
    \caption{\textbf{MCE results of different datasets for UV.}}\label{mle_result}
\end{table*}

Recall the statement we made in~\cref{sec:union_cut} that there are two types of UVs in UnionCut: 1) background UV: when the UV\textquotesingle s seed patch is a background patch, the UV is expected to return a binary mask indicating the background in the image; and 2) foreground UV: when a foreground patch is used as the seed patch of a UV, the UV outputs a binary mask indicating foreground regions in the image. To solve Inequality~\ref{ineq} with $P(s_i \in B)$ as the variable, four items in it need to be known in advance, \ie $P(p_j \in \hat{F}|s_i \in F, p_j \in B)$, $P(p_j \in \hat{F}|s_i \in B, p_j \in B)$, $P(p_j \in \hat{F}|s_i \in F, p_j \in F)$, and $P(p_j \in \hat{F}|s_i \in B, p_j \in F)$, with the meaning of each item as follows:\\
\textbf{1) }$\mathbf{P(p_j \in \hat{F}|s_i \in F, p_j \in B)}$: the probability of a background patch $p_j \in B$ being returned by a foreground UV with a foreground patch $s_i \in F$ as the seed patch;\\
\textbf{2) }$\mathbf{P(p_j \in \hat{F}|s_i \in B, p_j \in B)}$: the probability of a background patch $p_j \in B$ being returned by a background UV with a background patch $s_i \in B$ as the seed patch;\\
\textbf{3) }$\mathbf{P(p_j \in \hat{F}|s_i \in F, p_j \in F)}$: the probability of a foreground patch $p_j \in F$ being returned by a foreground UV with foreground patch $s_i \in F$ as the seed patch;\\
\textbf{4) }$\mathbf{P(p_j \in \hat{F}|s_i \in B, p_j \in F)}$: the probability of a foreground patch $p_j \in F$ being returned by a background UV with a background patch $s_i \in B$ as the seed patch.

These four items cannot be calculated directly. Therefore, we estimate their values statistically by MCE (Monte Carlo Estimation) with images in different datasets~\cite{duts, coco,ecssd}. For example, to estimate $P(p_j \in \hat{F}|s_i \in F, p_j \in B)$, the percentage of an image\textquotesingle s background patches being returned by each foreground UV is saved, and the mean value of these saved percentages across all images is calculated as the estimated $\hat{P}(p_j \in \hat{F}|s_i \in F, p_j \in B)$. Since $P=\{p_1, p_2, ...,p_{784}\}$ is defined as the set of patches of an image, a dataset $D$ with $K$ images can be denoted by $D=\{P_1, P_2, ...,P_K\}$. For the $k^{th}$ image in the dataset, its foreground patch set and background patch set given by ground truth are represented by $F_k$ and $B_k$, respectively. Besides, $\hat{F}_{kl}$ denotes the set of patches returned by a UV on the $k^{th}$ image with $s_l \in P_k$ as the seed patch. \cref{mle} provides the MCE solution to estimate $P(p_j \in \hat{F}|s_i \in F, p_j \in B)$, $P(p_j \in \hat{F}|s_i \in B, p_j \in B)$, $P(p_j \in \hat{F}|s_i \in F, p_j \in F)$, and $P(p_j \in \hat{F}|s_i \in B, p_j \in F)$, where $\mid \cdot \mid$ calculates the number of elements in a set.

\begin{equation} \label{mle}
\resizebox{8cm}{!}{$
\begin{aligned}
&\hat{P}(p_j \in \hat{F}|s_i \in F, p_j \in B) = \frac{\sum_{k=1}^{K}\sum_{s_l \in F_k}\frac{\mid \hat{F}_{kl} \cap B_k \mid}{\mid B_k \mid}}{\sum_{k=1}^{K}\mid F_k \mid}\\
&\hat{P}(p_j \in \hat{F}|s_i \in B, p_j \in B) = \frac{\sum_{k=1}^{K}\sum_{s_l \in B_k}\frac{\mid \hat{F}_{kl} \cap B_k \mid}{\mid B_k \mid}}{\sum_{k=1}^{K}\mid B_k \mid}\\
&\hat{P}(p_j \in \hat{F}|s_i \in F, p_j \in F) = \frac{\sum_{k=1}^{K}\sum_{s_l \in F_k}\frac{\mid \hat{F}_{kl} \cap F_k \mid}{\mid F_k \mid}}{\sum_{k=1}^{K}\mid F_k \mid}\\
&\hat{P}(p_j \in \hat{F}|s_i \in B, p_j \in F) = \frac{\sum_{k=1}^{K}\sum_{s_l \in B_k}\frac{\mid \hat{F}_{kl} \cap F_k \mid}{\mid F_k \mid}}{\sum_{k=1}^{K}\mid B_k \mid}
\end{aligned}
$}
\end{equation}

\cref{mle_result} illustrates the estimation of the four items in Inequality~\ref{ineq} with~\cref{mle} on different datasets. It can be observed that the probability (\ie $\hat{P}(p_j \in \hat{F}|s_i \in F, p_j \in B)$) of a background patch being returned by a foreground UV is very low, and vice versa (\ie $\hat{P}(p_j \in \hat{F}|s_i \in B, p_j \in F)$). However, there is a significant difference in the probability (\ie $\hat{P}(p_j \in \hat{F}|s_i \in F, p_j \in F)$) of foreground patches being returned by a foreground UV, compared to the probability (\ie $\hat{P}(p_j \in \hat{F}|s_i \in B, p_j \in B)$) of background patches being returned by a background UV. Specifically, when a background patch is selected, other background patches are more likely to be segmented, whereas when a foreground patch is selected, other foreground patches are not as easily segmented. In other words, during the voting process, background UVs are relatively active—they, while refusing to vote for foreground patches, nominate most of the background patches. In contrast, foreground UVs are relatively inactive: they refuse to vote for background patches, and also only nominate a small portion of the foreground patches. This leads to UnionCut’s aggregated final voting results showing that background patches receive significantly more votes than foreground patches, even in extreme cases—for instance, when the number of background UVs is smaller than that of foreground UVs (i.e., when the foreground occupies the majority of the image), this situation still holds. This difference between the two types of UVs is counterintuitive. but precisely, this critical distinction allows UnionCut to work robustly, even in images where the background occupies a tiny portion. 

Based on the above findings, we can explain why UnionCut works robustly for the detection of foreground unions with ensemble learning theories.

\textbf{UnionCut\textquotesingle s Effectiveness and Robustness} Assuming that we regard all UVs as weak classifiers for background classification (regardless of whether they are foreground UVs or background UVs), meaning that the regions with a value of 1 in all UVs\textquotesingle output binary masks are considered background, then a background UV can always correctly classify the background. For a foreground UV, it can be interpreted here as a background weak classifier handling hard cases: although it may erroneously return foreground as background, its less active nature and its tendency to suppress both foreground and background in its output mean that the errors it produces have minimal impact on the final voting result. Therefore, in the final aggregated result of UnionCut, the correct background segmentation produced by the background UVs dominates, causing UnionCut as a strong classifier to reliably output the background regions of the image, which, after being inverted and thresholded, become the final foreground union output by UnionCut.

\begin{table*}[t]
    \centering
    \footnotesize
    \begin{tabular}{lcccc}
        \toprule
        Dataset & $\hat{P}(p_j \in \hat{F}|s_i \in F, p_j \in B)$ & $\hat{P}(p_j \in \hat{F}|s_i \in B, p_j \in B)$ & $\hat{P}(p_j \in \hat{F}|s_i \in F, p_j \in F)$ & $\hat{P}(p_j \in \hat{F}|s_i \in B, p_j \in F)$\\
        \hline
        COCO2014 train~\cite{coco} & 0.536 & 0.8832 & 0.6554 & 0.3962\\
        COCO20K~\cite{coco,simeoni2021localizing} & 0.536 & 0.8832 & 0.6554 & 0.3977\\
        ECSSD~\cite{ecssd} & 0.3348 & 0.9467 & 0.7131 & 0.2668\\
        DUTS-TR~\cite{duts} & 0.3567 & 0.9568 & 0.6966 & 0.2853\\
        DUTS-TE~\cite{duts} & 0.3518 & 0.9183 & 0.7195 & 0.2476\\
        \bottomrule
    \end{tabular}
    \caption{\textbf{MCE results of different datasets for cosine similarity matching.}}\label{mce_result_cos}
\end{table*}

\textbf{Impact of the Proportion of the Background in the image on UnionCut} With MCE results in~\cref{mle_result}, Inequality~\ref{ineq} can be solved by substituting the estimated values in ~\cref{mle_result} into Inequality~\ref{ineq}. As shown in ~\cref{solution}, taking COCO2014 train~\cite{coco} as an example, Inequality~\ref{ineq} holds when $P(s_i \in B) > 0.1344$. Since $P(s_i \in B)$ in an image is approximated by the percentage of area occupied by background patches, $P(s_i \in B) > 0.1344$ can be understood as indicating that when the background occupies more than 13.44\% of the image area, UnionCut can robustly detect the foreground union in the image. In other words, even if there are large objects in an image, UnionCut works robustly if the foreground union occupies less than 86.56\% of the image.

\begin{table}[htbp]
    \centering
    \begin{tabular}{lc}
        \toprule
        Dataset & Solution\\
        \hline
        COCO2014 train~\cite{coco} & $P(s_i \in B) > 0.1344$\\
        COCO20K~\cite{coco,simeoni2021localizing} & $P(s_i \in B) > 0.1372$\\
        ECSSD~\cite{ecssd} & $P(s_i \in B) > 0.1862$\\
        DUTS-TR~\cite{duts} & $P(s_i \in B) > 0.1787$\\
        DUTS-TE~\cite{duts} & $P(s_i \in B) > 0.1967$\\
        \bottomrule
    \end{tabular}
    \caption{\textbf{Solutions of Inequality~\ref{ineq} for UV based on the MCE results in \cref{mle_result}}}\label{solution}
\end{table}

\section{Examples of Foreground Union Detection on Images with Large Foreground}
\label{large_foreground_example}
\begin{figure}[htbp]
    \centering
    \includegraphics[width=8cm]{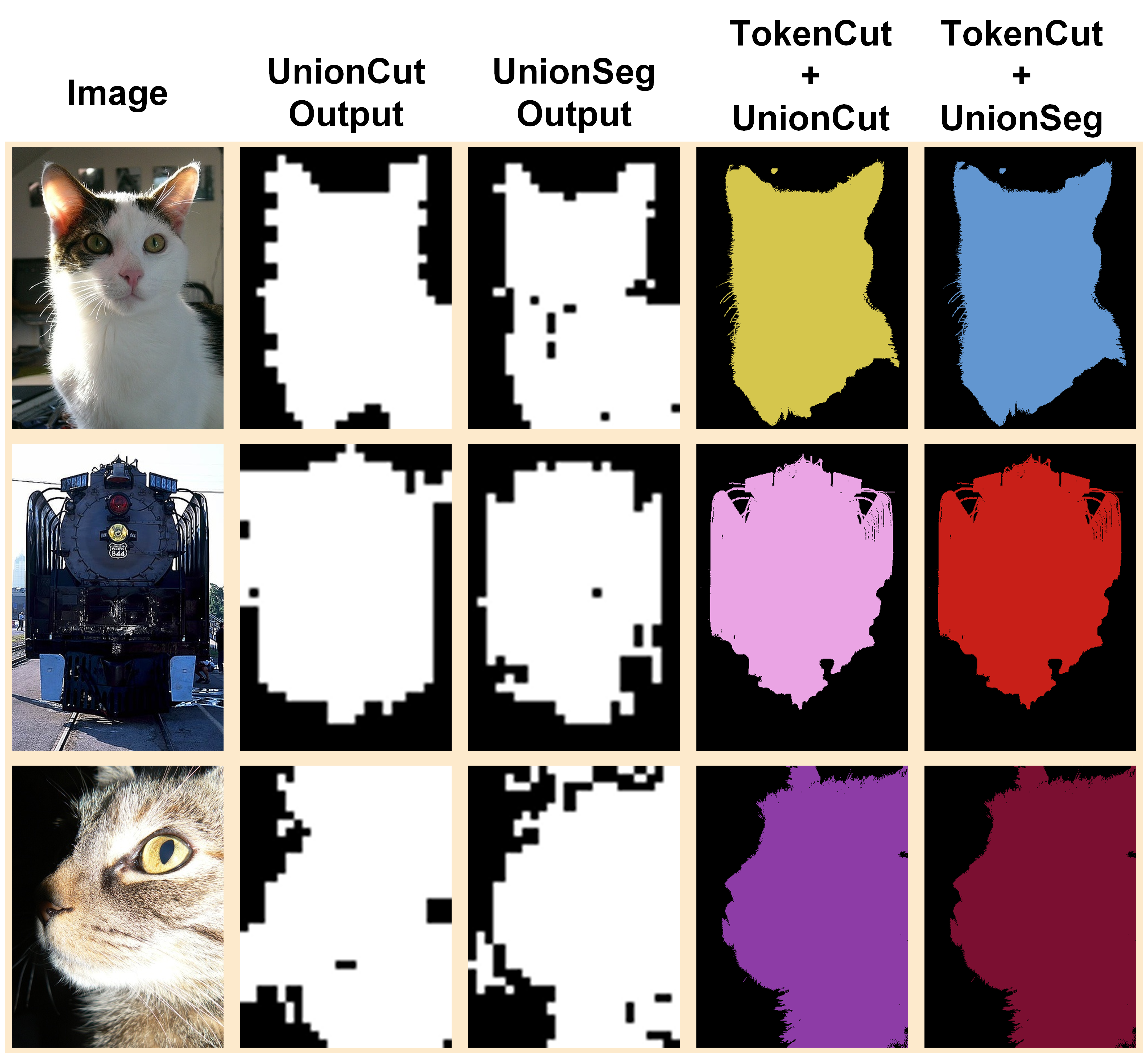}
    \caption{\textbf{Examples of UnionCut and UnionSeg\textquotesingle s foreground union detection results on images with large foreground}.}
    \label{vis3}
\end{figure}

In this section, we show UnionCut and UnionSeg\textquotesingle s successful examples on images whose background is smaller than the foreground to support our claim in~\cref{sec:union_cut} that UnionCut can stay effective on images of large foreground areas. As shown in~\cref{vis3}, the foreground union of three images occupied mainly by the foreground are accurately detected by both UnionCut and UnionSeg, demonstrating that UnionCut and UnionSeg stay effective on images of small backgrounds.

\begin{table}[htbp]
    \centering
    \begin{tabular}{lc}
        \toprule
        Dataset & Solution\\
        \hline
        COCO2014 train~\cite{coco} & $P(s_i \in B) > 0.1968$\\
        COCO20K~\cite{coco,simeoni2021localizing} & $P(s_i \in B) > 0.1973$\\
        ECSSD~\cite{ecssd} & $P(s_i \in B) > 0.3574$\\
        DUTS-TR~\cite{duts} & $P(s_i \in B) > 0.3361$\\
        DUTS-TE~\cite{duts} & $P(s_i \in B) > 0.3541$\\
        \bottomrule
    \end{tabular}
    \caption{\textbf{Solutions of Inequality~\ref{ineq} for consine similarity matching based on the MCE results in \cref{mce_result_cos}}}\label{baseline_solution}
\end{table}

\section{Comparison between UV and Straitforward Feature Matching}
\label{UBVvsCOS}
In this section, we continue to use the mathematical and statistical methods in Appendix~\ref{prove} to demonstrate the unique advantages and necessity of UV, compared with other straightforward feature matching methods. Here, we make matching similar patches with cosine similarity used by \cite{simeoni2021localizing, FOUND, MaskDistill} as an example. Specifically, given the seed patch $s_i \in P$ and its feature vector $k_i \in K$, the patches sharing features similar to $s_i$ can be obtained and defined as $\hat{F}=\{p_j|p_j \in P, k_j^Tk_i>0 \}$. Based on \cref{mle}, $P(p_j \in \hat{F}|s_i \in F, p_j \in B)$, $P(p_j \in \hat{F}|s_i \in B, p_j \in B)$, $P(p_j \in \hat{F}|s_i \in F, p_j \in F)$, and $P(p_j \in \hat{F}|s_i \in B, p_j \in F)$ of cosine similarity matching can be estimated, as shown in \cref{mce_result_cos}. Compared with UV, when a foreground patch is selected as the seed patch, the probability (\ie $P(p_j \in \hat{F}|s_i \in F, p_j \in B)$) of a background patch being returned is much larger, and vice versa, indicating that consine similarity feature matching is more likely to make mistakes as a weak classifier than UV. Besides, the difference between $P(p_j \in \hat{F}|s_i \in B, p_j \in B)$ and $P(p_j \in \hat{F}|s_i \in F, p_j \in F)$ is not significant, indicating that cosine similarity matching of a foreground seed patch and background seed patch are similarly active. This reduces the dominant influence of the cosine similarity of the background seed patches on the final aggregated voting results. 

Furthermore, we can also solve Inequality~\ref{ineq} with the estimated results in \cref{mce_result_cos}. As shown in \cref{baseline_solution}, taking DUTS-TE as an example, the aggregated strong classifier with cosine similarity matching as weak classifiers only works on images with the background area occupies over 35.41\% (compared with 19.67\% for UV), \ie the foreground should occupy less than 64.59\% areas of the image (compared with 80.33\% for UV), which much limits the robustness and effectiveness of the strong classifier since there are many image cases that break this constraint, examples can be seen in ~\cref{vis3}. 

\textbf{Weak Classifier\textquotesingle s Diversity} We can also compare UV and cosine similarity matching in terms of the requirement of ensemble methods with ensemble learning theories. Specifically, to obtain a robust enough strong classifier, there is a diversity requirement for weak classifiers, \ie weak classifiers should be different from each other and as diverse as possible~\cite{ensemble}. A large number of homogeneous weak classifiers can cause the performance of the aggregated strong classifier to degrade to that of a weak classifier. 

Compared to UV, using straightforward cosine similarity matching of a seed patch as a weak classifier leads to a large number of homogeneous weak classifiers. That is because directly computing the cosine similarity between a seed patch\textquotesingle s feature and those of other patches requires calculating the distance matrix of all patches, and this distance matrix is symmetric. This symmetry means that if a patch $p_j \in P$ appears in the cosine similarity matching results when using another patch $p_i \in P$ as the seed patch, then $p_i$ will also appear in the cosine similarity matching results when using $p_j$ as the seed patch. Ultimately, this results in a certain homogeneity in the output of the weak classifiers.

In contrast, UV is more diverse as a weak classifier because the output of each UV of a seed patch $s_i \in P$ depends not only on the features of $s_i$ but also on the features of its anti-seed patches $B_f=\{p_b|p_b \in P, b \neq f, k_b^Tk_f<0 \}$. Since the anti-seed patches for each seed patch $s_i$ are different, the introduction of anti-seed patches eliminates the previously mentioned symmetry, making UV as a weak classifier more varied, leading to UnionCut as a strong classifier being more robust.

In summary, based on the previous discussion with \cref{mle_result}, \cref{mce_result_cos}, \cref{solution} and \cref{baseline_solution}, our proposed UV has advantages in robustness over straightforward feature matching, and is less limited by the impact of large area occupied by the foreground. Moreover, although our UV is more complex than straightforward feature matching, its pipeline, which models the image as a graph, can effectively consider both the seed patch and the anti-seed patches during execution, which can hardly be done by straightforward feature matching. This makes our UV technically irreplaceable (this does not mean that no other algorithm can achieve the same effect as the UV, but more research is needed to explore this possibility).

\section{Reliability of Corner Prior}
\label{reliability_corner}
To explore the reliability of the corner prior used by UnionCut, we focus on its success rate on images from different datasets~\cite{coco, ecssd, duts, omron}. Specifically, we calculate the success rate by assessing the proportion of images in each dataset where the union of the ground truth occupies less than four corners of the image. As shown in \cref{successful_rate_corner}, the corner prior\textquotesingle s success rate achieves over 99\% on every selected dataset, indicating its robustness in checking the reliability of the foreground union mask detected by UnionCut. 

\begin{table}[htbp]
    \centering
    \begin{tabular}{lc}
        \toprule
        Dataset & Success Rate\\
        \hline
        COCO2014 train~\cite{coco} & 99.92\% \\
        COCO2017 train~\cite{coco} & 99.92\% \\
        COCO20K~\cite{coco,simeoni2021localizing} & 99.92\% \\
        ECSSD~\cite{ecssd} & 100\% \\
        DUTS-TR~\cite{duts} & 99.83\% \\
        DUTS-TE~\cite{duts} & 99.89\% \\
        DUTS-OMRON~\cite{omron} & 100\% \\
        \bottomrule
    \end{tabular}
    \caption{\textbf{Successful rate of the corner prior utilized by UnionCut with different datasets.}}\label{successful_rate_corner}
\end{table}

\begin{figure*}[htbp]
    \centering
    \includegraphics[width=15cm]{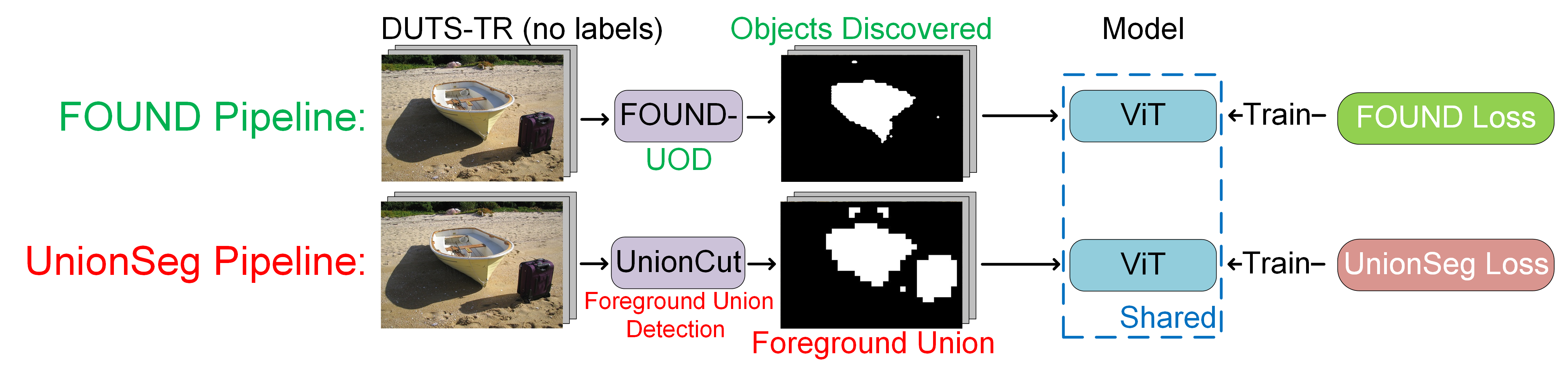}
    \caption{The comparison of the framework between FOUND~\cite{FOUND} and UnionSeg.}
    \label{found_vs_unionseg_pipeline}
\end{figure*}

\begin{figure*}[htbp]
    \centering
    \begin{subfigure}[b]{0.3\linewidth}
        \centering
        \includegraphics[width=\linewidth]{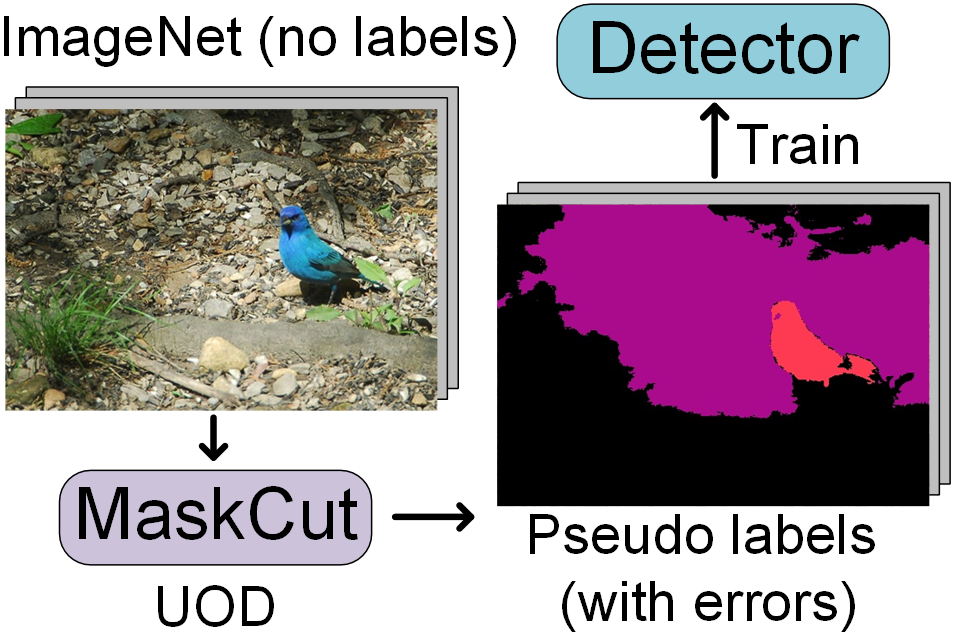}
        \caption{Pipeline of CutLER}
        \label{cutler_pipeline}
    \end{subfigure}
    \hspace{20pt} 
    \begin{subfigure}[b]{0.5\linewidth}
        \centering
        \includegraphics[width=\linewidth]{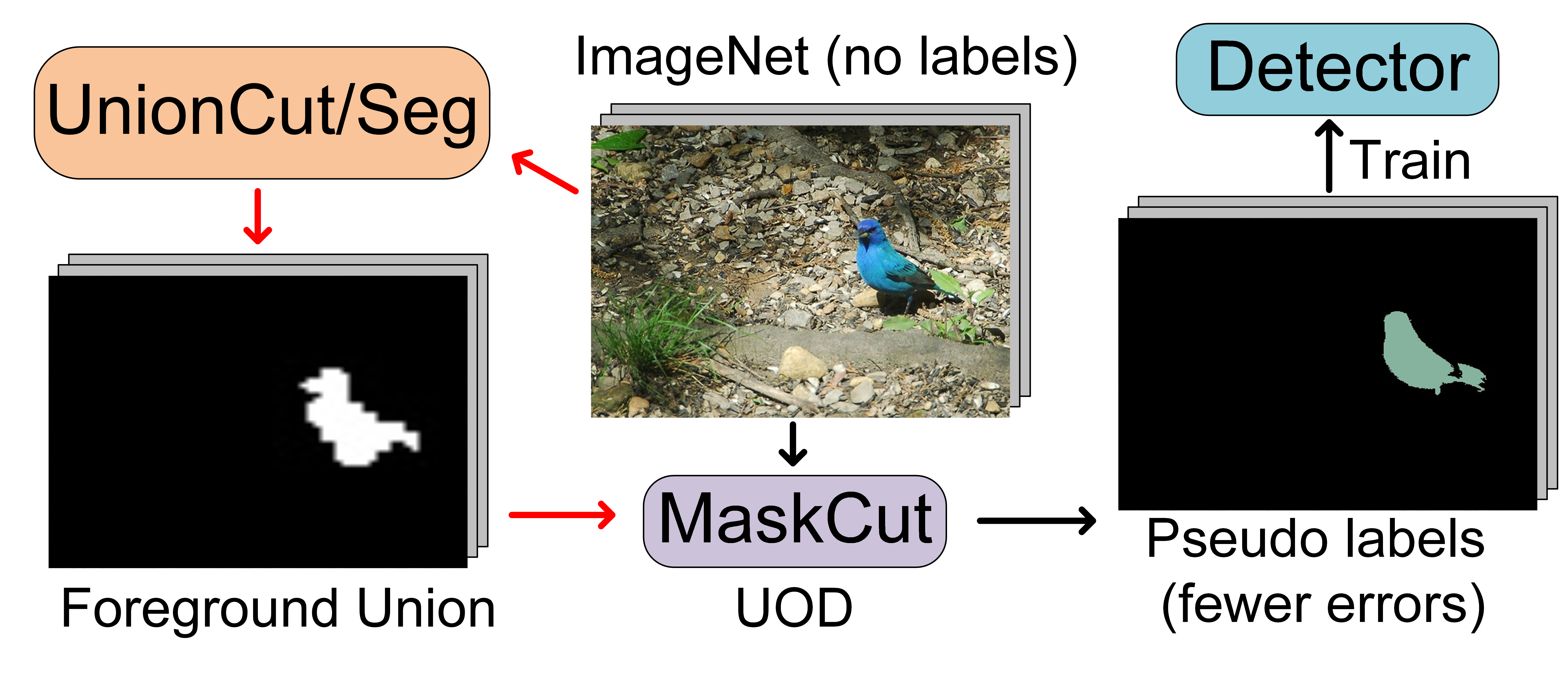}
        \caption{Pipeline of CutLER+UnionCut/UnionSeg}
        \label{cutler_pipeline_uc_us}
    \end{subfigure}
    \caption{The comparison between the pipeline of CutLER and CutLER+UnionCut/UnionSeg.}
\end{figure*}

\section{FOUND vs. UnionSeg}\label{UnionSeg_vs_FOUND}
\cref{found_vs_unionseg_pipeline} compares the pipeline of FOUND~\cite{FOUND} and UnionSeg. In~\cite{FOUND}, an unnamed UOD method, here referred to as ``FOUND-", is proposed to conduct UOD on the dataset DUTS-TR~\cite{duts}. After that, the object discovered by FOUND- is used as pseudo-labels to train a ViT surrogate model (\ie FOUND). Similarities and differences between FOUND and UnionSeg are listed below:\\
\textbf{Similarities:} they share the same model structure.\\
\textbf{Differences:}\\
1) their loss functions and training settings are different;\\
2) Their pseudo-labels have different sources and represent different meanings: the pseudo-labels for FOUND are generated by FOUND‐ and represent the objects discovered in the image (which do not necessarily cover the majority of objects in the image). In contrast, UnionSeg\textquotesingle s pseudo-labels are generated by UnionCut and are designed to cover most of the object regions in the image, i.e., the foreground union;\\
3) Based on the difference mentioned above, the function of UnionSeg and FOUND are also different: FOUND and FOUND- are proposed for unsupervised object discovery (\ie UOD), while UnionSeg and UnionCut are utilized to predict the union of foreground area (\ie foreground union) in an image, which are used as foreground priors for UOD methods.

\section{Combining UnionCut/UnionSeg with Existing UOD methods and Implementation Details}
\label{implementation}
The core idea of combining UnionCut or UnionSeg with existing UOD methods is to replace their default foreground priors with UnionCut/UnionSeg. In this section, we introduce how to apply UnionCut/UnionSeg to existing UOD methods. Note that the methods designed by us are not the only way to combine UnionCut/UnionSeg with existing UOD methods. We welcome the proposal of more advanced combining methods to further enhance UnionCut/UnionSeg\textquotesingle s boosting effect on UOD algorithms. More technical details of this section can be seen in our code.

\subsection{Basic Usage of UnionCut and UnionSeg}\label{basic_usage}
UnionCut and UnionSeg can be integrated into UOD algorithms in various ways; here, we show their basic usage. Assuming the foreground union $U$ detected by UnionCut or UnionSeg as the ground truth, the precision of a mask discovered by UOD algorithms can be calculated by~\cref{precision}
\begin{equation} \label{precision}
\resizebox{5cm}{!}{$
Precision(mask, U) = \frac{Area(mask \cap U)}{Area(mask)}
$}
\end{equation}
where $Area(\cdot)$ returns the area of a mask, based on which UnionCut and UnionSeg can be utilized in two ways:\\
1) For judging whether a discovered area belongs to the foreground, an area will be considered as part of the foreground in the image if its precision is higher than a predetermined threshold $\theta$ (\eg 0.5);\\
2) To determine when to stop further discovery, exploration stops if the majority (a percentage $\gamma$, \eg 80\% area) of the foreground union given by UnionCut or UnionSeg has been discovered. 

In practice, we recommend setting $\theta=0.5$ and $\gamma=0.8$ when using the basic usage of UnionCut or UnionSeg.




\subsection{LOST+UnionCut/UnionSeg}
LOST~\cite{simeoni2021localizing}\textquotesingle s default foreground prior is as follows: An image is divided into patches corresponding to the serialized input of a ViT. The image is organized into a graph with each patch as a node. For any two patches in the graph, they are connected by links if their cosine similarity is more than 0. Then all patches are sorted by their degrees (\ie the number of links connected to a patch) in ascending order, and the first patch after being sorted is made as the foreground seed based on the assumption made by Sim{\'e}oni \etal that the area occupied by the foreground should be smaller than the background~\cite{simeoni2021localizing}. Then a discovered object region is expanded from the foreground seed across the entire image. More details can be found in \cite{simeoni2021localizing}.

In this paper, we combine LOST and UnionCut/UnionSeg by limiting the foreground seed selection range within the foreground union output by UnionCut/UnionSeg, \ie a patch will not be chosen as the seed if it is not in the foreground union, even though its degree is minimal. Besides, the discovered object area is expanded from the seed across the region of the foreground union instead of the entire image. 

\subsection{TokenCut+UnionCut/UnionSeg}
TokenCut is a single-object discovery method that divides an image into an object area and background (may contain other objects). To combine TokenCut with UnionCut/UnionSeg, we replace its default foreground prior with our approach, selecting the area with higher precision (~\cref{precision}) as the discovered object from the bipartition.

\subsection{FOUND+UnionCut/UnionSeg}
As introduced in \cref{UnionSeg_vs_FOUND}, FOUND is trained based on the output of FOUND- as pseudo-lables. We combine UnionSeg/UnionCut with FOUND by making the FOUND trained based on the intersection between FOUND-\textquotesingle s output and foreground union by UnionSeg/UnionCut as the pseudo-labels.

\subsection{CutLER+UnionCut/UnionSeg}
\subsubsection{The pipeline of combining CutLER and UnionCut or UnionSeg}
CutLER detects multiple objects in three steps, as shown in \cref{cutler_pipeline}: 1) MaskCut~\cite{wang2023cut} (a UOD method) discovers objects up to three times per image to generate pseudo-annotations on ImageNet~\cite{imagenet}; 2) training a Cascade Mask-RCNN~\cite{cascade} with these pseudo-annotations; 3) using the trained Cascade Mask-RCNN to update pseudo-annotations and retraining the model. As depicted in \cref{cutler_pipeline_uc_us}, we use UnionCut/UniongSeg to eliminate errors from MaskCut, enhancing the detector\textquotesingle s performance after training. Specifically, UnionCut/UniongSeg is integrated with CutLER at its first step,~\ie MaskCut. Without a foreground prior, MaskCut\textquotesingle s excessive discovery leads to misidentifying the background as the foreground, so Wang~\etal~\cite{wang2023cut} limits MaskCut to three discoveries per image. Using UnionCut/UniongSeg, we can remove this balance and make MaskCut stop when 80\% of the foreground union given by UnionCut/UniongSeg is detected, discarding discovered areas with precision (\cref{precision}) below 0.5. In the paper, all results related to CutLER and CutLER+UnionCut/UniongSeg are performances of the Cascade Mask-RCNN after training.

\subsubsection{Training Details of CutLER+UnionSeg}
We use the official implementation of CutLER~\cite{wang2023cut} in our experiments, using our implementation of MaskCut~\cite{wang2023cut} with UnionSeg to replace the initial MaskCut in the original CutLER. Following CutLER, a ViT-B/8~\cite{vit} pretrained by DINO~\cite{dino} is used as the image feature extractor for MaskCut. In terms of training CutLER+UnionSeg, first, we make our MaskCut+UnionSeg to conduct object discovery on all images (1.28 million) from ImageNet. Due to the large number of images to be processed, although our algorithm can judge when to stop discovery, we additionally made MaskCut+UnionSeg conduct up to 5 times exploration per image. With 18 processes running simultaneously, it took us 3 weeks to generate UOD results for all images, which were used as pseudo-annotations of instance segmentation for the dataset. Then, we used ImageNet and these pseudo-annotations to train a Cascade Mask-RCNN model with the official implementation of CutLER for 10,000 iterations with a learning rate of 0.01 and weight decay of 0.001 as a warm-up. After that, the trained model was used to update pseudo-annotations for images in ImageNet. We further trained the model for 60,000 external iterations with the updated pseudo-annotations, using a learning rate of 0.005 and weight decay of 0.0001. All other settings,~\eg optimizer, batch size and DropLoss threshold, were inherited entirely from the official CutLER.

\section{Performance Upper Boundary of UOD Methods Designed for Multiple Objects Discovery on Single Object Discovery}
\label{most}
Single object discovery requires UOD algorithms to predict only one bounding box for an image, and CorLoc is calculated by checking if the predicted bounding box matches any one object\textquotesingle s annotation in the image. However, when applying UOD methods able to predict multiple objects per image to single object discovery, strategies of selecting only one predicted bounding box are required to filter out redundant predictions, \eg selecting the predicted bounding box of the largest area or highest confidence. However, various strategies lead to different performances~\cite{FOUND}. As such, Rambhatla~\etal~\cite{most} proposed to use average best overlap~\cite{ss} to evaluate the performance upper boundary of UOD algorithms designed for multiple object discovery on single object discovery. Specifically, each predicted bounding box is compared with all ground truth bounding boxes in an image. If any prediction matches a ground truth bounding box, the UOD algorithm will be considered successful on this image. The essence of this approach is the assumption that an optimal strategy exists to select the predicted box from a set of predictions which is most likely to match the ground truth. Applying this assumed strategy allows for measuring the UOD algorithm\textquotesingle s performance on single object discovery in an ideal scenario, \ie its upper boundary performance. We also try to apply this strategy to evaluate the upper-boundary performance of CutLER (with or without UnionSeg) on single object discovery. As shown in \cref{corloc_ub}, after considering the upper boundary, the performance\textquotesingle s upper boundary of CutLER is also increased by UnionSeg and achieves state-of-the-art performance on all three benchmarks, indicating the effectiveness of our proposed UnionSeg.

\begin{table}[htbp]
    \centering
    \scriptsize
    \renewcommand\tabcolsep{3pt}
    \begin{tabular}{lcccc}
        \toprule
        Method & UB &VOC07 & VOC12 & COCO20K\\
        \hline
        \multicolumn{5}{c}{\textbf{- No learning -}}\\
        Selective Search~\cite{ss} & & 18.8 & 20.9 & 16.0\\
        EdgeBoxes~\cite{edgeboxes} & & 31.1 & 31.6 & 28.8\\
        Kim~\etal~\cite{kim} & & 43.9 & 46.4 & 35.1\\
        Zhang~\etal~\cite{zhang} & & 46.2 & 50.5 & 34.8\\
        DDT+~\cite{DDT} & & 50.2 & 53.1 & 38.2\\
        rOSD~\cite{rOSD} & & 54.5 & 55.3 & 48.5\\
        LOD~\cite{LOD} & & 53.6 & 55.1 & 48.5\\
        DINO-seg~\cite{dino,simeoni2021localizing}(ViT-S/16~\cite{dino}) & & 45.8 & 46.2 & 42.0\\
        LOST~\cite{simeoni2021localizing}(ViT-S/16~\cite{dino}) & & 61.9 & 64.0 & 50.7\\
        DSS~\cite{DSS}(ViT-S/16~\cite{dino}) & & 62.7 & 66.4 & 52.2\\
        MOST~\cite{most} & \Checkmark & 74.8 & 77.4 & 67.1\\
        \hdashline
        TokenCut~\cite{wang2022tokencut}(ViT-S/16~\cite{dino}) & & 68.8 & 72.1 & 58.8\\
        \textbf{TokenCut(ViT-S/16~\cite{dino})+UnionCut} & & 69.2\textbf{(0.4$\uparrow$)} & 72.3\textbf{(0.2$\uparrow$)} & 62.1\textbf{(3.3$\uparrow$)}\\
        \textbf{TokenCut(ViT-S/16~\cite{dino})+UnionSeg} & & 69.7\textbf{(0.9$\uparrow$)} & 72.7\textbf{(0.6$\uparrow$)} & 62.6\textbf{(3.8$\uparrow$)}\\
        \hline
        \multicolumn{5}{c}{\textbf{- With learning -}}\\
        FreeSOLO~\cite{freesolo,FOUND} & & 44.0 & 49.7 & 35.2\\
        LOD+CAD~\cite{simeoni2021localizing, wang2022tokencut} & & 56.3 & 61.6 & 52.7\\
        rOSD+CAD~\cite{simeoni2021localizing, wang2022tokencut} & & 58.3 & 62.3 & 53.0\\
        LOST+CAD~\cite{simeoni2021localizing}(ViT-S/16~\cite{dino}) & & 65.7 & 70.4 & 57.5\\
        SelfMask~\cite{selfmask} & & 72.3 & 75.3 & 62.7\\
        FOUND~\cite{FOUND}(ViT-S/16~\cite{dino}) & & 72.5 & 76.1 & 62.9\\
        \hdashline
        CutLER~\cite{wang2023cut}(ViT-B/8~\cite{dino})$\dagger$ & & 73.3 & 69.5 & 70.7\\
        \textbf{CutLER(ViT-B/8~\cite{dino})+UnionSeg} & & 73.8\textbf{(0.5$\uparrow$)} & 71.2\textbf{(1.7$\uparrow$)} & 72.4\textbf{(1.7$\uparrow$)}\\
        CutLER~\cite{wang2023cut}(ViT-B/8~\cite{dino})$\dagger$ & \Checkmark & 87.3 & 84.3 & 89.3\\
        \textbf{CutLER(ViT-B/8~\cite{dino})+UnionSeg} & \Checkmark & 87.4\textbf{(0.1$\uparrow$)} & 85.1\textbf{(0.8$\uparrow$)} & 89.3\\
        \bottomrule
    \end{tabular}
    \caption{\textbf{Performances of UOD methods on single object discovery, with CorLoc as the metric and the upper boundary of UOD methods detecting multiple objects considered.} UB: the short for Upper Boundary. CAD: a class-agnostic detector trained with the output of no-learning UOD methods as the ground truth. $\dagger$: results of official implementations and checkpoints by our measurement.}\label{corloc_ub}
\end{table}

\section{UnionCut\textquotesingle s Effectiveness on Self-supervised Instance Segmentation}
\label{unioncut_instance}
In this section, CutLER (MaskCut) and TokenCut are combined with UnionCut before being used to generate pseudo-labels for images in the dataset. After that, these pseudo-labels are used to train a class-agnostic SOLOv2~\cite{solov2} model. The subset (2913 images) of VOC2012~\cite{voc} where images are with pixel-level annotations are used as the benchmark for training and evaluating the performance of the model~\cite{solov2} trained with pseudo-labels given by different UOD algorithms. Five folds are used for cross-validation. Specifically, for each fold, the dataset is divided into training, validation, and test sets in a ratio of 8:1:1. The model is trained with images of the training part and corresponding pseudo-labels generated by a UOD method. The model\textquotesingle s weights are saved for every 1000 iterations. The validation set and handcrafted ground truth provided by VOC2012 are used to evaluate the model\textquotesingle s weights of each iteration. The weights that perform best on the validation set are selected, whose performance is evaluated with the test set with handmade ground truth and reported here. All training in this experiment uses a fixed random seed of $3407$ and trains the model for $30,000$ iterations per fold. Models\textquotesingle ~parameters are updated using an Adam optimizer~\cite{adam} with a learning rate $0.0001$ and a mini-batch size $16$. Average precision (AP) is used as the metric. Consistent experimental setups apply to all other UOD methods. \cref{unioncut_instance_table} shows that UnionCut improves the quality of pseudo-labels given by MaskCut and TokenCut, indicating its effectiveness in boosting the performance of UOD methods as a robust foreground prior.

\begin{table}[t]
    \centering
    \small
    \renewcommand\tabcolsep{5pt}
    \renewcommand\arraystretch{1}
    \begin{tabular}{lcc}
        \toprule
        Method & $AP_{0.5}^{mask}$ & $AP_{0.5}^{box}$\\
        \hline
        LOST$\dagger$ & 10.2 & 14.9\\
        MaskDistill$\ddagger$ & 11.3 & 16.2\\
        \hdashline
        TokenCut$\dagger$ & 19.0 & 22.8\\
        TokenCut+UnionCut & \textbf{19.2(0.2$\uparrow$)} & \textbf{22.9(0.1$\uparrow$)}\\
        CutLER$\dagger$ & 18.5 & 22.0\\
        CutLER+UnionCut & \textbf{20.9(2.4$\uparrow$)} & \textbf{23.5(1.5$\uparrow$)}\\
        \bottomrule
    \end{tabular}
    \caption{\textbf{Performance reported on VOC12 of SOLOv2\cite{solov2} trained with pseudo-labels given by different UOD methods.} $\dagger$: codes from the official implementation. $\ddagger$: our replication based on the paper.}\label{unioncut_instance_table}
\end{table}


Note that unlike \cref{instance_all} in~\cref{subsec:ssis} where ImageNet (1.28 million images) is used as the training set, considering the computational cost of UnionCut, only VOC12 (2983 images with pixel-level annotations) is utilized for training the model in this experiment, which is much fewer than ImageNet. Since UnionSeg\textquotesingle s effectiveness has been discussed in~\cref{subsec:ssis} in terms of instance segmentation, this section aims to show the effectiveness of UnionCut additionally and is not for making the model achieve state-of-the-art performance.

\section{Examples of Images without Foreground Union Fully Annotated}
\label{sec:voc_example}
\begin{figure}[htbp]
    \centering
    \includegraphics[width=7.5cm]{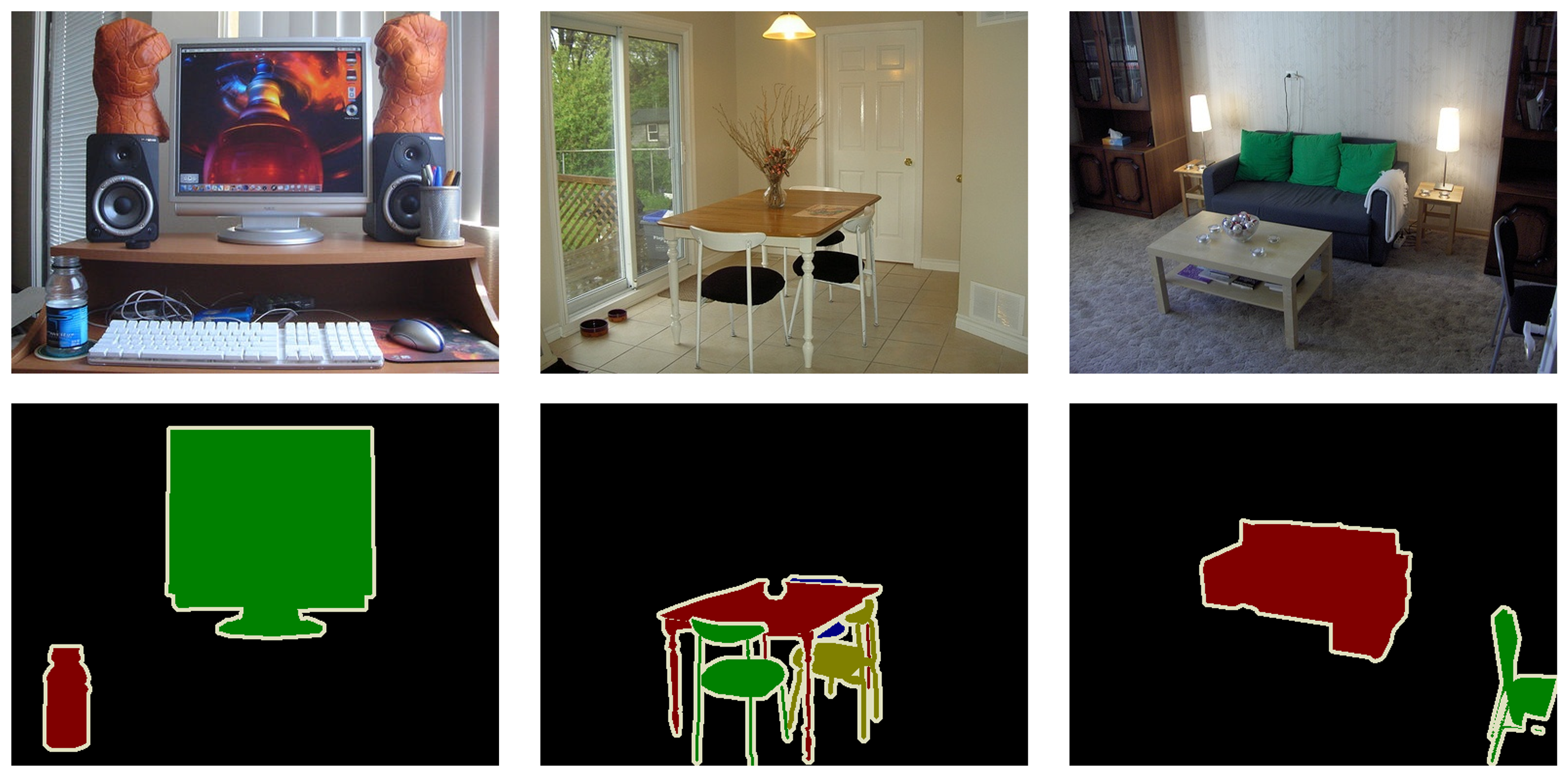}
    \caption{\textbf{Visualization of images in VOC12 whose ground truth union do not cover the foreground union of the image.}}
    \label{voc_example}
\end{figure}

This section provides example images from VOC12, where the union of the ground truth does not fully cover the foreground union. As shown in \cref{voc_example}, the keyboard, stereo, cup, etc., are not labelled by the ground truth (the left column); the vase and light are not labelled (the mid column); and the end table is not labelled (the right column).

\section{Qualitative Analysis}
\label{more_vis}
In this section, we visualize and analyse how UnionCut and UnionSeg boost the performance of our selected baseline UOD algorithms,~\ie TokenCut and MaskCut, and provide more visualization.

\begin{figure}[htbp]
    \centering
    \includegraphics[width=8cm]{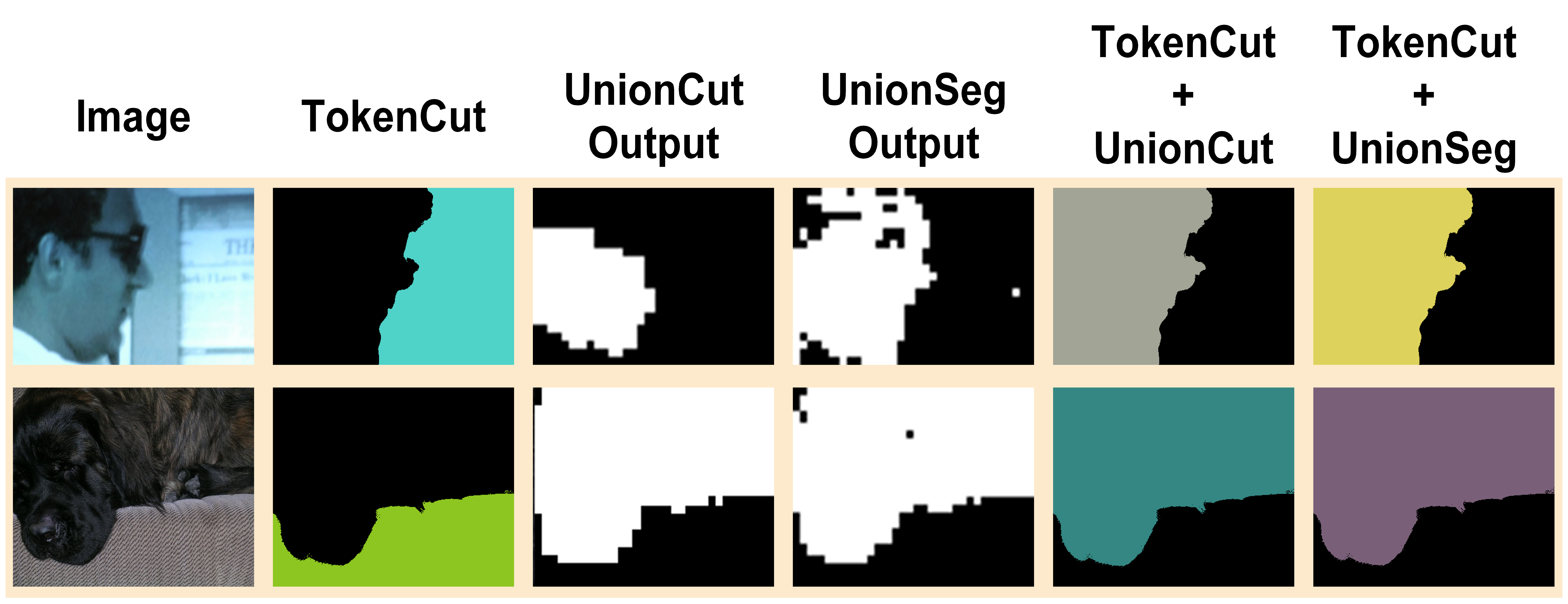}
    \caption{TokenCut\textquotesingle s errors fixed by UnionCut and UnionSeg.}
    \label{vis1}
\end{figure}

\textbf{Foreground Judgement} UnionCut/UnionSeg enables UOD methods to judge if a discovered area belongs to the foreground. Taking TokenCut as an example, as shown in \cref{vis1}, TokenCut initially chooses the background part from the bipartition as the discovered result. With UnionCut or UnionSeg, the segmentation mostly covered by the foreground union is chosen to correct TokenCut\textquotesingle s error.

\begin{figure}[htbp]
    \centering
    \includegraphics[width=8cm]{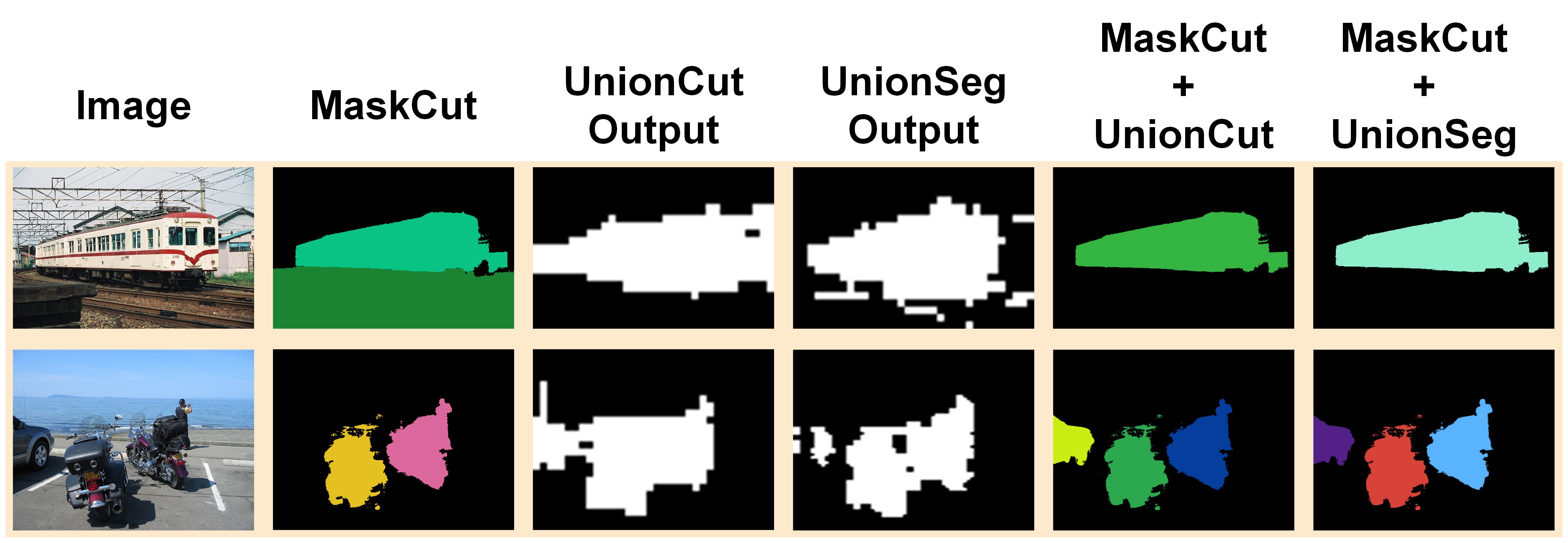}
    \caption{Examples of removing MaskCut\textquotesingle s detection errors and making its discovery stop at appropriate time.}
    \label{vis2}
\end{figure}

\textbf{Complete Discovery} UnionCut/UnionSeg enables UOD algorithms, especially those detecting multiple objects, to judge when to stop further discovery without under or over-discovery. \cref{vis2} illustrates two examples where UnionCut and UnionSeg help MaskCut remove a misidentified background area (the 1st row) and prevent missing one object (the car on the left of the 2nd image) by ensuring the majority of the foreground union is discovered.

\begin{figure*}[htbp]
    \centering
    \includegraphics[width=17cm]{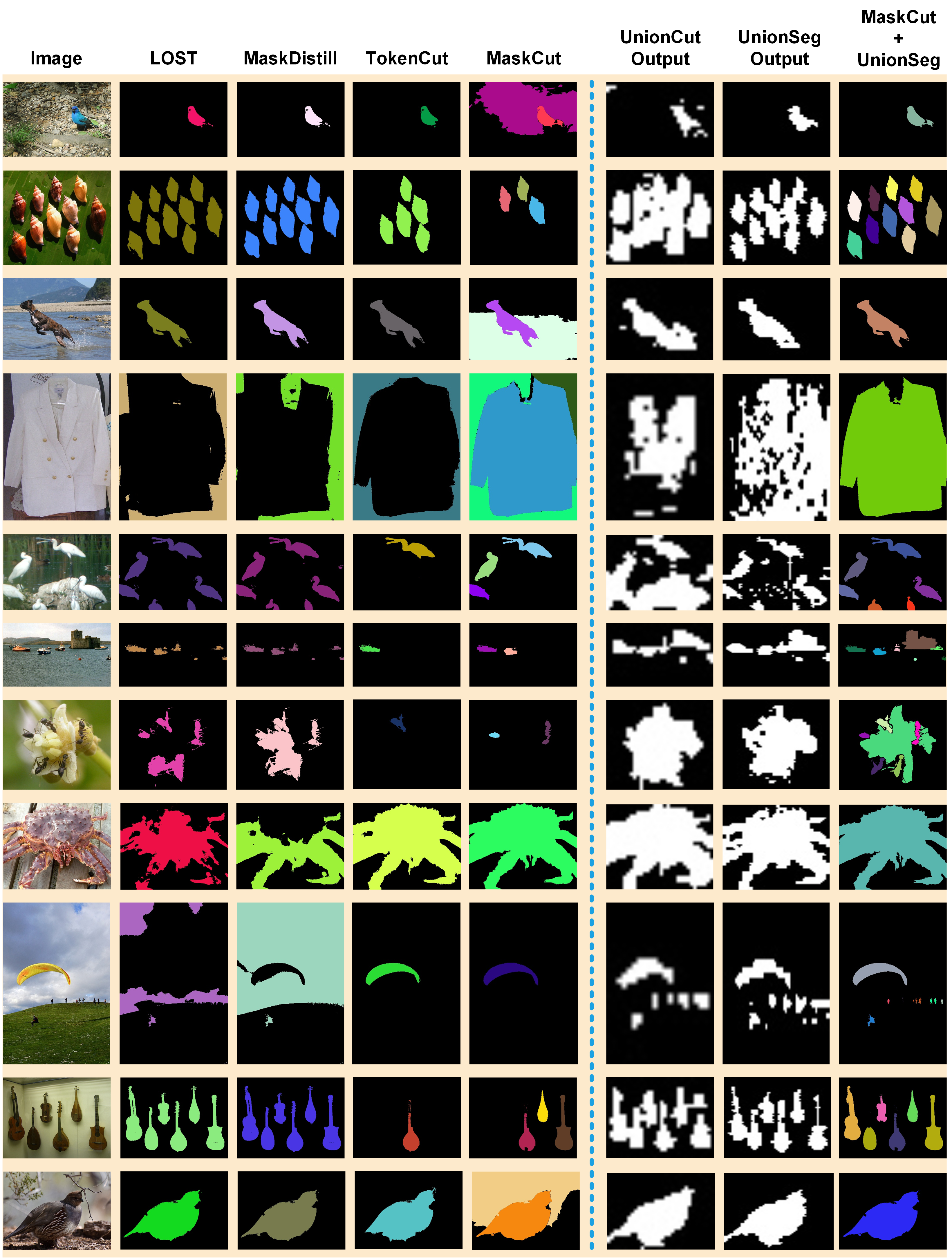}
    \caption{\textbf{Visualization of results given by different UOD algorithms.} Masks of the same colour are the result of a one-time discovery.}
    \label{additional_vis}
\end{figure*}

\textbf{Additional Visualization} \cref{additional_vis} provides additional qualitative results of different UOD methods, indicating that UnionCut and UnionSeg detect foreground union accurately, fix errors made by MaskCut without missing objects or including background, and, as such, improve the performance of MaskCut. Only MaskCut here (with or without UnionSeg) conducts discovery multiple times, while others conduct UOD one time per image. We made MaskCut conduct up to 3 times discovery per image following the recommendation of the original work~\cite{wang2023cut}. As for our MaskCut+UnionSeg, to show that UnionSeg enables UOD algorithms to stop discovery at the appropriate time, we made MaskCut+UnionSeg conduct up to 50 times discovery per image.

\end{document}